\documentclass[mlmain,onecolumn]{jmlr}

\jmlrvolume{}
\firstpageno{1}
\editors{List of editors' names}

\jmlrproceedings{PMLR}{Symmetry and Geometry in Neural Representations Workshop 2025}
\jmlryear{2025}
\jmlrworkshop{Symmetry and Geometry in Neural Representations}

\title[Exact Learning Dynamics of ICL in Linear Transformers]{Exact Learning Dynamics of In-Context Learning in Linear Transformers and its application}

\author{%
\Name{Nischal Mainali} \Email{nischal.mainali@mail.huji.ac.il}\\
\addr The Edmond and Lily Safra Center for Brain Sciences, Hebrew University of Jerusalem
\AND
\Name{Lucas Teixeira} \Email{}\\
\addr Principles of Intelligent Behavior in Biological and Social Systems (PIBBSS)
}

\begin{document}

\maketitle

\begin{abstract}
Transformer models exhibit remarkable in-context learning (ICL), adapting to novel tasks from examples within their context, yet the underlying mechanisms remain largely mysterious. Here, we provide an exact analytical characterization of ICL emergence by deriving the closed-form stochastic gradient descent (SGD) dynamics for a simplified linear transformer performing regression tasks. Our analysis reveals key properties: (1) a natural separation of timescales directly governed by the input data's covariance structure, leading to staged learning; (2) an exact description of how ICL develops, including fixed points corresponding to learned algorithms and conservation laws constraining the dynamics; and (3) surprisingly nonlinear learning behavior despite the model's linearity. We hypothesize this phenomenology extends to non-linear models. To test this, we introduce theory-inspired macroscopic measures (spectral rank dynamics, subspace stability) and use them to provide mechanistic explanations for (1) the sudden emergence of ICL in attention-only networks and (2) delayed generalization (grokking) in modular arithmetic models. Our work offers an exact dynamical model for ICL and theoretically grounded tools for analyzing complex transformer training.
\end{abstract}


\section{Introduction}

The transformer architecture~\citep{vaswani2017attention} has revolutionized machine learning, particularly through its capability for in-context learning (ICL). ICL allows models to adapt their behavior based on examples provided within the input context, obviating the need for task-specific parameter updates \citep{brown2020language}. This phenomenon links conceptually to meta-learning, characterized by distinct timescales: slow learning via weight updates during training and fast adaptation based on context during inference \citep{vonOswald2023transformers}. Unlike traditional supervised learning, ICL enables generalization to novel tasks by conditioning on contextual examples \citep{chen2021evaluating}. Mechanistically, attention layers can implement adaptive processing analogous to fast weights \citep{schmidhuber1992learning, schlag2021linear, vonOswald2023transformers}, suggesting that transformers might internally simulate learning algorithms \citep{akyurek2023context, vonOswald2023transformers, ahn2023transformers}. Despite compelling empirical demonstrations, a full theoretical grasp of how transformers achieve ICL remains unclear.

To address this challenge, our work presents an in-depth analysis of a simplified, mathematically tractable transformer model performing in-context linear regression. Our model incorporates specific simplifications—linear attention \citep{vonOswald2023transformers, akyurek2023context}, a structured input embedding \citep{garg2022can, akyurek2023context}, mean-squared error loss, and a continuous-time approximation of learning dynamics—yet retains core aspects of ICL. This deliberate simplification allows us to analytically trace how contextual information modulates the model's predictions. By leveraging methods from the study of exactly solvable models, introduced first in deep linear networks \citep{saxe2014exact}, we derive a precise characterization of the learning dynamics under stochastic gradient descent (SGD). We analyze the resulting phenomenology, including the emergence of multiple learning timescales dictated by data statistics \citep{saxe2014exact, rahaman2019spectral}. This detailed understanding of a simplified system aims to provide basic insights applicable to the interpretation and analysis of more complex transformer models exhibiting ICL.
In particular, we believe that this detailed understanding can inform interpretability efforts and evaluation methodologies for more complex, real-world models and argue the point by quantitatively analyzing the properties of model parameters in 1) simple attention only transformer model that exhibit sudden emergence of in context learning during training time, and 2) grokking model that exhibits delayed generalization.

\subsection{Related Work}

Transformers exhibit remarkable in-context learning (ICL), adapting to new tasks using context examples without explicit parameter updates \citep{brown2020language}. This connects to meta-learning's dual timescales: slow weight learning and fast context adaptation \citep{vonOswald2023transformers}. Early concepts like fast weights \citep{schmidhuber1992learning} suggested context-based adaptation mechanisms. Recent studies demonstrate that linear self-attention layers can be mathematically equivalent to fast weight update rules \citep{schlag2021linear, vonOswald2023transformers}, enabling Transformers to implement learning algorithms within their forward pass \citep{vonOswald2023transformers}.

Several studies confirm that Transformers can encode standard learning algorithms. \citet{akyurek2023context} proved by construction that Transformers can implement gradient descent (GD) and ridge regression, finding empirically that trained models mimic these algorithms and converge towards Bayesian optimal predictors for linear tasks. Similarly, \citet{vonOswald2023transformers} demonstrated that Transformers trained on regression problems learn GD internally, with attention updates mapping directly to GD steps, effectively becoming "mesa-optimizers". In parallel work \citet{ahn2023transformers}, and \citet{zhang} provided theoretical guarantees that standard training yields preconditioned GD at the fixed-point of the training dynamics across a range of setups. These findings collectively suggest that standard training procedures can induce Transformers to implicitly learn optimization algorithms. Extending this, \citet{fu2024transformers} found evidence that Transformers might learn higher-order methods resembling Newton's method, explaining superior performance on ill-conditioned problems. Alternative perspectives conceptualize ICL as Bayesian inference, where models infer latent task variables from prompts \citep{xie2022explanation}, or as kernel regression, where attention patterns reflect similarity-based weighting akin to kernel methods \citep{han2023explaining}. These views provide explanations for observed phenomena like prompt sensitivity and the utility of relevant examples \citep{xie2022explanation, han2023explaining}. The precise mechanism—gradient-based, Bayesian, or kernel-like—likely depends on model scale, architecture, and training data \citep{akyurek2023context, vonOswald2023transformers, han2023explaining, xie2022explanation, ahn2023transformers, zhang}.

At the same time, analyzing learning dynamics has a rich history, notably through the study of deep linear networks \citep{saxe2014exact}. \citet{saxe2014exact} derived exact solutions for these models, revealing nonlinear dynamics and multiple timescales despite linearity. Learning often occurs in stages, marked by plateaus followed by rapid error reduction as different eigenmodes of the data correlation are sequentially learned. Learning timescales are typically eigenvalue-dependent, with dominant modes learned faster \citep{saxe2014exact, advani2017high}. In nonlinear networks, exact solutions are rare, but phenomena like spectral low rank bias indicate structured learning dynamics: models tend to learn low-frequency (smooth) function components before high-frequency details \citep{rahaman2019spectral}. This bias is linked to the Neural Tangent Kernel (NTK) spectrum, where larger eigenvalues associated with smoother functions lead to faster learning \citep{rahaman2019spectral, Jacot2018Neural}, potentially explaining the generalization capabilities of overparameterized models.

Separation of timescales emerges as a common theme across linear dynamics \citep{saxe2014exact}, nonlinear spectral bias \citep{rahaman2019spectral}, meta-learning \citep{vonOswald2023transformers} and the abrupt emergence of ICL or induction heads \citep{olsson2022context} hints at similar timescale separation. Indeed, there have been theoretical  results in non-linear transformer establishing distinct phases in the learning dynamics \cite{boix-adsera, chen2024} further supporting the idea. Alternative models like associative memories instead connect ICL to retrieval dynamics \citep{ramsauer2021hopfield}.

Most recently, parallel studies have also explored the theoretical properties of linear attention. \citet{lu2025} and \citet{lyu2025} similarly identify that linear attention models learn to whiten or precondition data; however, these works focus primarily on asymptotic behavior and scaling laws in multi-task regimes, respectively. Additionally, \citet{zhang2025} analyze the exact learning dynamics for scalar outputs ($\mathbb{R}^d \to \mathbb{R}^1$) within multi-head architectures and different QK parametrization. In contrast, our work provides an non-asymptotic closed-form characterization of the learning dynamics for vector-valued ($\mathbb{R}^d \to \mathbb{R}^d$) regression and apply the insights from the theory for interpreting non-linear models.

\section{Model Setup}
\label{sec:model_setup_main_paper}

\subsection{Linear Transformer Architecture}
\label{subsec:linear_transformer_architecture_main_paper}

We analyze a single-layer transformer with a single linear attention head. In contrast to standard architectures using softmax attention \citep{vaswani2017attention}, we employ simplifications common in theoretical studies \citep{vonOswald2023transformers, akyurek2023context}. The model utilizes two weight matrices, $W^Q \in \mathbb{R}^{D \times D}$ (query/key) and $W^P \in \mathbb{R}^{D \times D}$ (output/projection), and replaces the attention non-linearity with the identity function ($\phi(x)=x$). The layer's computation is defined as (see Appendix~\ref{sec:model_architecture}-\ref{sec:linear_transformer}, Eq.~\eqref{eq:transformer_output}):
\begin{equation}
    \label{eq:linear_attention_main_paper}
    f(Z) = Z + W^{P} \left( \frac{Z Z^{\top}}{N} \right) W^{Q} Z.
\end{equation}
Here, $Z \in \mathbb{R}^{D \times (N+1)}$ is the input sequence embedding, and $N+1$ is the sequence length. The learnable parameters are $\Theta = \{ W^P, W^Q \}$. This linear attention model, while simplified, is sufficient for learning in-context linear regression \citep{vonOswald2023transformers}.

Relevant parameter submatrices are defined based on the input embedding structure (Appendix~\ref{sec:model_variables}). We partition $W^Q$ such that its first $d$ columns form $q = \left[  q_1^T \, \mid \, q_2^T \right]^T$, and $W^P$ such that its bottom $d$ rows form $p^T = \left[ p_1 \, \mid \, p_2 \right]$. All submatrices $q_1, q_2, p_1, p_2$ are in $\mathbb{R}^{d \times d}$. More precisely, we have the following parametrization:
\begin{align}
W^P &= \begin{pmatrix} \cdot & \cdot \\ p_1 & p_2 \end{pmatrix} \quad \text{with} \quad p^T = \begin{pmatrix} p_1 & p_2 \end{pmatrix}, \\
W^Q &= \begin{pmatrix} q_1 & \cdot \\ q_2 & \cdot \end{pmatrix} \quad \text{with} \quad q = \begin{pmatrix} q_1 \\ q_2 \end{pmatrix},
\end{align}

\subsection{Embedding Structure and Task Setup for In-Context Regression}
\label{subsec:embedding_task_setup_main_paper}

The model is trained to perform in-context linear regression. Given a prompt containing $N$ example pairs $(x_i, y_i) \in \mathbb{R}^{d} \times \mathbb{R}^d$, it must predict the output $y_q$ for a query input $x_q \in \mathbb{R}^d$ based on the linear relationship demonstrated in the prompt. This setup extends the $\mathbb{R}^{d} \to \mathbb{R}^{1}$ tasks studied in some prior work \citep{akyurek2023context, garg2022can} to the more general vector-valued regression setting $\mathbb{R}^{d} \to \mathbb{R}^{d}$.

We employ a structured embedding strategy commonly used in ICL studies \citep{akyurek2023context, garg2022can, raffel2020exploring}. For an input sequence $(x_1, y_1, \dots, x_N, y_N, x_q)$, the embedding matrix $Z \in \mathbb{R}^{D \times (N+1)}$ (with $D=2d$) is constructed as:
\begin{equation}
    \label{eq:embedding_main_paper}
    Z = \begin{pmatrix}
        x_1 & \dots & x_N & x_q \\
        y_1 & \dots & y_N & \mathbf{0}_d
    \end{pmatrix}.
\end{equation}
This embedding structure (Appendix~\ref{sec:model_variables}, Eq.~\eqref{eq:input_matrix_Z}) confers permutation invariance over context examples via the $Z Z^T$ term, distinguishes the query input using the zero vector padding, and implicitly encodes the input-output pairing structure without requiring explicit positional encodings.

\subsection{Task Statistics and Training Procedure}
\label{subsec:task_statistics_training_main_paper}

We define a generative process for the linear regression tasks. Input data $x$ are sampled from a zero-mean Gaussian distribution $\mathcal{N}(0, \Sigma_x)$. The covariance matrix $\Sigma_x$ is assumed diagonalizable by an orthogonal matrix $U$, such that $\Sigma_x = U S U^{\top}$, where $S = \text{diag}(s_1, \dots, s_d)$ contains the eigenvalues. The corresponding output is generated via $y = W x$, where $W$ is a task-specific matrix.

We generate a fixed set of $P$ task matrices $\{W^\mu\}_{\mu=1}^P$ prior to training. Each task matrix is constructed as $W^\mu = U \Lambda^\mu U^{\top}$, where $\Lambda^\mu = \text{diag}(\lambda_{\mu,1}, \dots, \lambda_{\mu,d})$ has eigenvalues sampled independently from $\mathcal{N}(0, 1)$. Crucially, the *same* orthogonal matrix $U$ is used for both $\Sigma_x$ and $W^\mu$. This alignment ensures that all relevant matrices commute in the eigenbasis of $\Sigma_x$, significantly simplifying the theoretical analysis (see Appendix~\ref{sec:W_matrix_generation} and \ref{sec:learning_parameters}).

The model's prediction $\hat{y}$ for query $x_q$ under task $\mu$ (derived from the $(N+1)$-th column of $f(Z^\mu)$) is evaluated against the ground truth $y_q^\mu = W^\mu x_q$ using the Mean Squared Error (MSE) loss function:
\begin{align}
    \label{eq:loss_function_main_paper}
  \mathcal{L}_{\mu} \left( \Theta \right) &= \frac{1}{2} \| \hat{y} - y_q^{\mu} \|_2^2.
\end{align}
The use of MSE is a simplification for analytical tractability compared to typical auto-regressive losses.

Training employs SGD with a small learning rate $\eta$. An epoch involves presenting all $P$ tasks, each time sampling fresh data contexts $\{x_i, x_q\}$. Parameter updates occur after processing each task sequentially. The small $\eta$ assumption justifies approximating the discrete learning process with continuous-time dynamics (see Appendix~\ref{sec:learning_parameters}).

\section{Theory}
\label{sec:main_results}
\begin{figure*}[t!]
    \centering
    \includegraphics[width=0.65\linewidth]{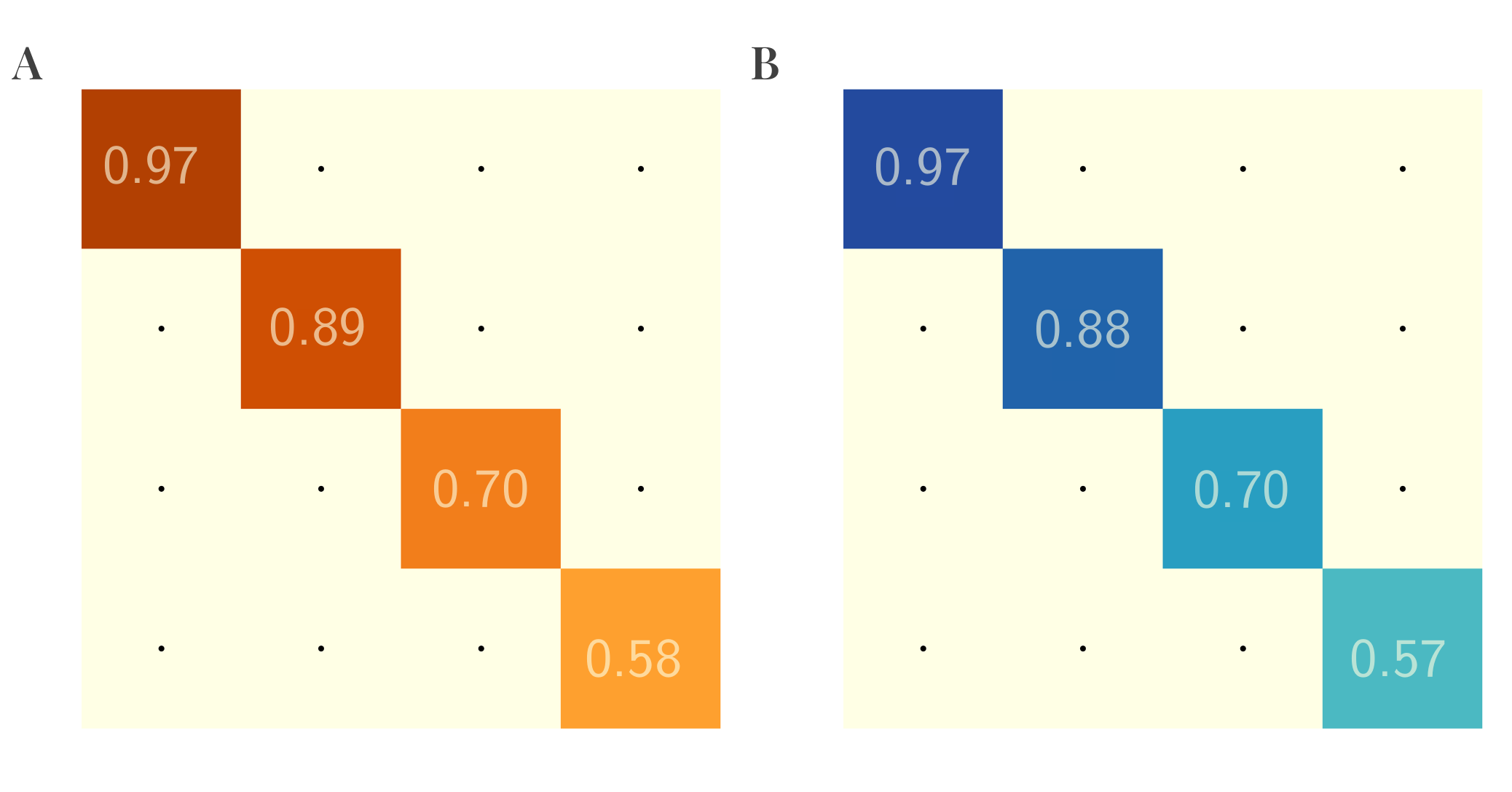}
    \caption{Fixed point parameters: theory vs. simulation. (A) Learned parameters ($p_2 q_1$) in batch-averaged simulation, diagonalized in the input covariance basis. (B) theoretical prediction for the fixed point $p_2 q_1 (\infty)$, showing precise quantitative agreement.}
    \label{fig:Figure1}
\end{figure*}
\textbf{Simplified predictor.} The model predicts the query output $\hat{y} \in \mathbb{R}^d$ using the bottom $d$ elements of the final output column $[f(Z)]_{:, N+1}$. As derived in Appendix~\ref{sec:simplified_predictor} (Eq.~\eqref{eq:predictor_output_final}), this prediction simplifies to a linear transformation of the query input:
\begin{equation}
    \label{eq:predictor_output_simplified_main_paper}
    \hat{y} = \left( p^{T} \hat{\Gamma} q \right) x_q,
\end{equation}
where $\hat{\Gamma} = \frac{Z Z^{\top}}{N}$ is the empirical covariance of the embedded sequence. The matrix $p^{T} \hat{\Gamma} q$ effectively encodes the linear function inferred from the context. The MSE loss for task $\mu$ (Eq.~\ref{eq:loss_function_main_paper}) uses this predictor with the task-specific covariance $\hat{\Gamma}^\mu$:
\begin{align}
    \label{eq:loss_function_main_paper_simple}
  \mathcal{L}_{\mu} \left( \Theta \right) &= \frac{1}{2} \left\| \left( p^{T} \hat{\Gamma}^{\mu} q \right) x_{q} - W^{\mu} x_{q} \right\|_2^2.
\end{align}

We proceed by analyzing the learning dynamics in the continuous-time limit, averaging over the quenched randomness of tasks $W^\mu$ and the annealed randomness of data samples $x_i, x_q$.

\textbf{Null Parameter Dynamics and Predictor Simplification.} A key simplification arises from identifying $p_1$ and $q_2$ as "null parameters." Unlike some prior theoretical works which simplify the optimization landscape by assuming specific parameterizations (e.g., rank constraints or symmetric weight assumptions) a priori, our simplification is derived dynamically. We show in Appendix~\ref{sec:null_parameters} that if initialized at zero, the expected gradients for $p_1$ and $q_2$ vanish identically \cite{zhang,ahn2023transformers}. This implies that the dynamics naturally confine the optimization to the relevant subspace without enforcing hard constraints, theoretically justifying fixing $p_1 = \mathbf{0}$ and $q_2 = \mathbf{0}$. Combined with a large-$N$ approximation for the empirical covariance $\hat{\Gamma}^\mu$ (which neglects the query's contribution, see Appendix~\ref{sec:y_data_generation}, Eq.~\eqref{eq:approx_empirical_covariance_Gamma_mu_block}), the predictor for task $\mu$ takes a remarkably simpler form (derived in Appendix~\ref{sec:null_parameters}, Eq.~\eqref{eq:predictor_output_task_mu_null_params}):
\begin{equation}
    \label{eq:predictor_simplified_final_main}
    \hat{y}_{\mu} \approx p_2 W^{\mu} \hat{\Sigma}_{x} q_1 x_q,
\end{equation}
where $\hat{\Sigma}_x = \frac{1}{N} \sum_{i=1}^N x_i x_i^\top$. This simplified predictor, involving only parameters $p_2$ and $q_1$, forms the foundation for our subsequent analysis of learning dynamics.
Empirically, even with small non-zero initialization, the learning dynamics drives $p_1$ and $q_2$ towards zero.
\begin{figure*}[t!]
    \centering
    \includegraphics[width=0.9\linewidth]{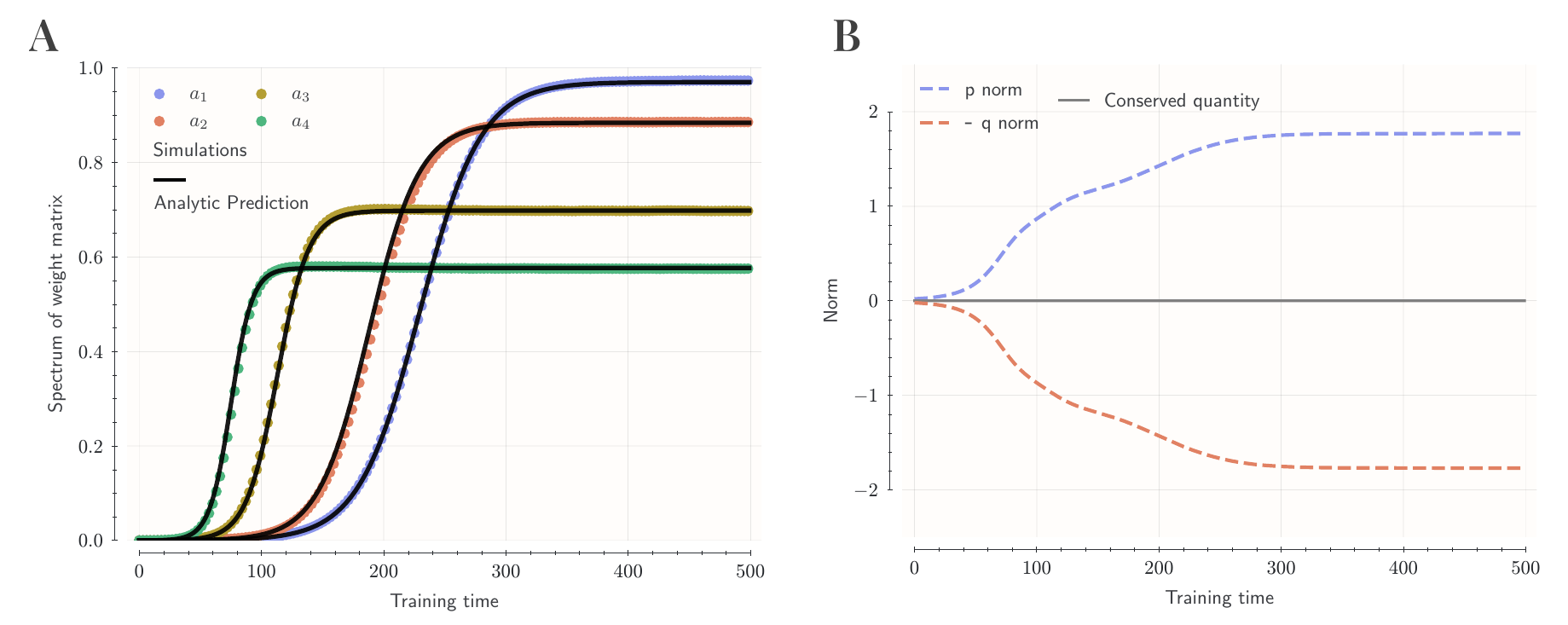}
    \caption{Parameter dynamics: theory vs. simulation. (A) Empirical, batch-averaged, evolution of the diagonal elements of the learned parameter $\bar{a}(t) = \bar{p}_2(t) \bar{q}_1(t)$ compared with theoretical predictions (Eq.~\ref{eq:a_alpha_solution_main}), showing excellent agreement. (B) Mirror evolution of two terms in conserved quantity and confirmation of the stability of the conserved quantity $\mathcal{C} = \lVert \bar{p}_2 \rVert_F^2 - \lVert \bar{q}_1 \rVert_F^2$.}
    \label{fig:Figure2}
\end{figure*}
\textbf{Non-linear Learning Dynamics and Spectral Decoupling.} To analyze the dynamics of $p_2$ and $q_1$, we assume spectral alignment: parameters $p_2(t), q_1(t)$ diagonalize in the same basis $U$ as the data covariance $\Sigma_x$, i.e., $p_2(t) = U \bar{p}_2(t) U^{\top}$ and $q_1(t) = U \bar{q}_1(t) U^{\top}$, where $\bar{p}_2, \bar{q}_1$ are diagonal (Appendix~\ref{sec:learning_parameters}, Eqs.~\eqref{eq:p2_decomposition}-\eqref{eq:q1_decomposition}). Averaging over stochasticity and taking the continuous-time limit yields coupled ODEs for the diagonal parameter matrices (Appendix~\ref{sec:learning_parameters}, Eqs.~\eqref{eq:dp2_dt_continuous}-\eqref{eq:dq1_dt_continuous}):
\begin{align}
    \frac{\mathrm{d} \bar{p}_2}{\mathrm{d}t} &= - \eta P S^2 \left( \bar{p}_2 \bar{q}_1^2 s^\infty(S) - \bar{q}_1 \right) \label{eq:ode_p2bar_main} \\
    \frac{\mathrm{d} \bar{q}_1}{\mathrm{d}t} &= - \eta P S^2 \left( \bar{q}_1 \bar{p}_2^2 s^\infty(S) - \bar{p}_2 \right), \label{eq:ode_q1bar_main}
\end{align}
where $s^\infty(S) = (\frac{N+1}{N}S + \frac{\text{Tr}(S)}{N}I)$ is a diagonal matrix encoding the influence of context length $N$ and data statistics $S$. These equations reveal fundamentally non-linear dynamics, despite the linearity of the attention mechanism itself. The learning rates are explicitly modulated by the squared eigenvalues $S^2$, indicating faster learning along directions of higher input variance. The product structure of terms like $\bar{p}_2 \bar{q}_1^2$ leads to cooperative growth from small initial values.

The system converges to a fixed point where $\frac{\mathrm{d}\bar{p}_2}{\mathrm{d}t} = \mathbf{0}$ and $\frac{\mathrm{d}\bar{q}_1}{\mathrm{d}t} = \mathbf{0}$. This occurs when the product of the parameters reaches $\bar{p}_2(\infty) \bar{q}_1(\infty) = [s^\infty(S)]^{-1}$. This fixed point result precisely matches those found in fixed-point analyses of linear transformers trained for in-context regression \citep{ahn2023transformers, zhang}. Figure~\ref{fig:Figure1} confirms this theoretical prediction against simulations. The effective linear transformation learned by the model at convergence is derived by substituting this fixed point into the predictor (Eq.~\ref{eq:predictor_simplified_final_main}):
\begin{equation}
    \label{eq:learned_algorithm_main}
    \hat{y}_{\mu}(\infty) \approx W^{\mu} \left[ \hat{\Sigma}_x \left( \frac{N+1}{N}\Sigma_x + \frac{\text{Tr}(\Sigma_x)}{N}I \right)^{-1} \right] x_q.
\end{equation}
In the large context limit ($N \to \infty$), assuming $\hat{\Sigma}_x \to \Sigma_x$, the term in brackets approaches $I$, yielding $\hat{y}_{\mu}(\infty) \to W^\mu x_q$. This demonstrates that the model learns to perfectly implement the target linear map. The term in brackets acts as a form of data-dependent preconditioning, related to ridge regression or preconditioned gradient descent updates identified in other works \citep{ahn2023transformers, vonOswald2023transformers}.

\textbf{Conserved Quantity and Reduced Degrees of Freedom.} The ODE system (Eqs.~\ref{eq:ode_p2bar_main}-\ref{eq:ode_q1bar_main}) exhibits a conserved quantity: $\mathcal{C} = \lVert \bar{p}_2 \rVert_F^2 - \lVert \bar{q}_1 \rVert_F^2$ remains constant throughout training (Figure~\ref{fig:Figure2}B, Appendix~\ref{sec:dynamical_equations_conserved_quantity}). This conservation law stems from an underlying scaling symmetry in the loss function and implies that the parameters evolve on a lower-dimensional manifold within the full parameter space, constraining their trajectory. This suggests a form of implicit regularization or low-dimensional learning, reminiscent of phenomena observed in other deep learning contexts \citep{biroli2022kernel, gerace2020disentangling}.
The existence of this conserved quantity can also be derived in a more principled way by observing that the loss function \(\mathcal{L} \left( \bar{p}_2 , \bar{q}_1\right)\) is invariant under \(\bar{p}_2 \to \eta \bar{p}_2\) and \(\bar{q}_1 \to \frac{\bar{q}_1}{\eta}\), and using Noether's theorem.

\textbf{Analytical Solution for Learning Trajectories and Timescales.} The dynamics decouple along the eigenmodes $\alpha = 1, \dots, d$. Defining the timescale $\tau_\alpha = (\eta P s_\alpha^2)^{-1}$ and the mode-specific fixed point inverse $s_\alpha^\infty = \frac{(N+1)s_\alpha + \operatorname{Tr}(S)}{N}$, the ODEs for the diagonal elements $p_\alpha, q_\alpha$ become (Appendix~\ref{sec:dynamical_equations_conserved_quantity}, Eqs.~\eqref{eq:dp_alpha_dt_timescale}-\eqref{eq:dq_alpha_dt_timescale}):
\begin{align}
    \label{eq:decoupled_odes_main}
    \tau_\alpha\frac{\mathrm{d}p_\alpha}{\mathrm{d}t} = q_\alpha\left(1 - p_\alpha q_\alpha s_\alpha^\infty \right); \quad
    \tau_\alpha\frac{\mathrm{d}q_\alpha}{\mathrm{d}t} = p_\alpha\left(1 - p_\alpha q_\alpha s_\alpha^\infty \right).
\end{align}
If initialized symmetrically ($p_\alpha(0) = q_\alpha(0)$), then $p_\alpha(t) = q_\alpha(t)$ holds for all time. Defining the product $a_\alpha(t) = p_\alpha(t) q_\alpha(t) (= p_\alpha(t)^2)$, its dynamics follow a logistic equation: $\tau_\alpha \frac{\mathrm{d}a_\alpha}{\mathrm{d}t} = 2a_\alpha(1 - a_\alpha s_\alpha^\infty)$. This equation admits an exact analytical solution (Figure~\ref{fig:Figure2}A, Appendix~\ref{sec:dynamical_equations_conserved_quantity}, Eq.~\eqref{eq:solution_a_alpha_t}):
\begin{equation}
    \label{eq:a_alpha_solution_main}
    a_\alpha(t) =  a_\alpha^\infty \cdot \frac{a_\alpha^0}{a_\alpha^0 + (a_\alpha^\infty - a_\alpha^0)\exp(-\frac{2t}{\tau_\alpha})},
\end{equation}
where $a_\alpha^\infty = 1/s_\alpha^\infty$ is the stable fixed point for mode $\alpha$, and $a_\alpha^0$ is the initial value.
The characteristic time $t_{\alpha}^{*}$ required to transition from a small initialization $\epsilon$ to near its fixed point is approximately $t_{\alpha}^{*} \approx \frac{1}{2\lambda P s_{\alpha}^2}\log \left( \frac{1}{s_{\alpha}^{\infty}\epsilon} \right)$.
This inverse dependence on $s_\alpha^2$ establishes timescale separation: modes corresponding to larger input eigenvalues $s_\alpha$ learn significantly faster than those associated with smaller eigenvalues, mirroring the multi-stage learning observed in deep linear networks \citep{saxe2014exact}.
\begin{figure}[t!]
    \centering
    \includegraphics[width=0.8\linewidth]{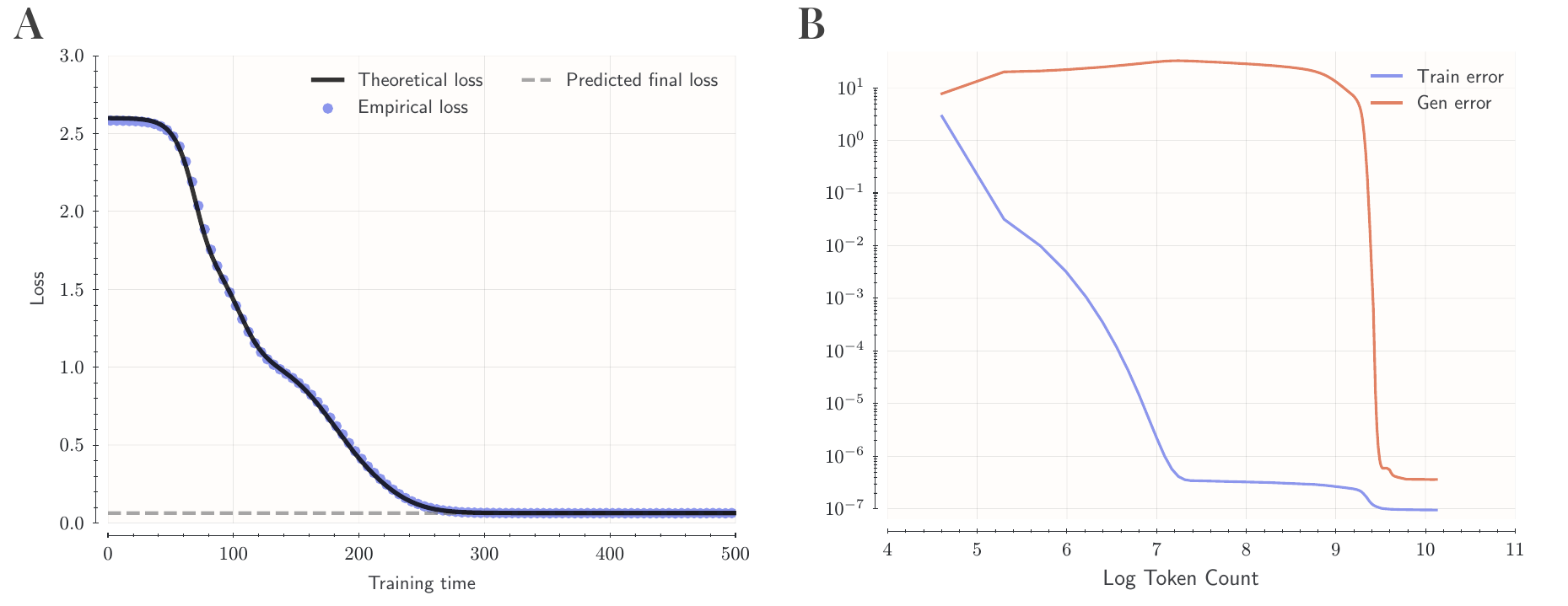}
    \caption{(A) Loss dynamics: theory vs. simulation. The empirical training loss curve closely matches the analytical prediction (Eq.~\ref{eq:loss_dynamics_main}), capturing characteristic plateaus and subsequent rapid decreases (cliffs) corresponding to the sequential learning of different spectral modes. (B) Training loss dynamics of a transformer model trained in modular arithmetic task displays qualitative similarity to the training dynamics of linear transformer in (A), and exhibits delayed generalized phenomena called ``grokking''. }
    \label{fig:Figure3}
\end{figure}
\textbf{Loss Dynamics and Plateaus.} Substituting the exact parameter trajectories $a_\alpha(t)$ into the expected loss yields the analytical loss evolution (Figure~\ref{fig:Figure3}, Appendix~\ref{sec:loss_analysis}, Eq.~\eqref{eq:expected_loss_final_a_alpha_inf}):
\begin{equation}
    \label{eq:loss_dynamics_main}
    \mathcal{L}(t) = \frac{1}{2} \sum_{\alpha=1}^d s_{\alpha} \left( \frac{s_{\alpha}}{a_{\alpha}^{\infty}} a_\alpha(t)^2 - 2s_\alpha a_\alpha(t) + 1 \right).
\end{equation}
This analytical form precisely captures the shape of the loss curve observed in simulations. The timescale separation inherent in $a_\alpha(t)$ directly translates into structured loss dynamics. The curve often exhibits plateaus, corresponding to periods where faster modes have converged but slower modes are still near their initial values (or slowly escaping saddle points near zero). Subsequent rapid drops ("cliffs") occur when these slower modes eventually learn, significantly reducing the remaining error (Figure~\ref{fig:Figure3}A). This phenomenology of transformer training dynamics is widely observed empirically, and we show a particular example from a ``grokking'' Transformer \citep{nanda2023grokking} in Figure~\ref{fig:Figure3}B, that has qualitative similarity to loss dynamics in our simplified model.
The initial loss decreases linearly, with a rate proportional to initial parameter values and input covariance eigenvalues.  The converged loss at $t \to \infty$ is non-zero due to the inherent randomness of finite context window and is given by $\mathcal{L}(\infty) = \frac{1}{2} \sum_\alpha \frac{s_\alpha (s_\alpha + \operatorname{Tr}(S))}{(N+1)s_\alpha + \operatorname{Tr}(S)}$.  The loss dynamics near convergence are dominated by exponential relaxation, governed by the slowest learning modes.
The final converged loss $\mathcal{L}(\infty)$ remains non-zero due to finite-$N$ effects related to noise in the empirical covariance estimates (Appendix~\ref{sec:loss_analysis}, Eq.~\eqref{eq:loss_infinity}).
\begin{figure*}[t!]
    \centering
    \includegraphics[width=\linewidth]{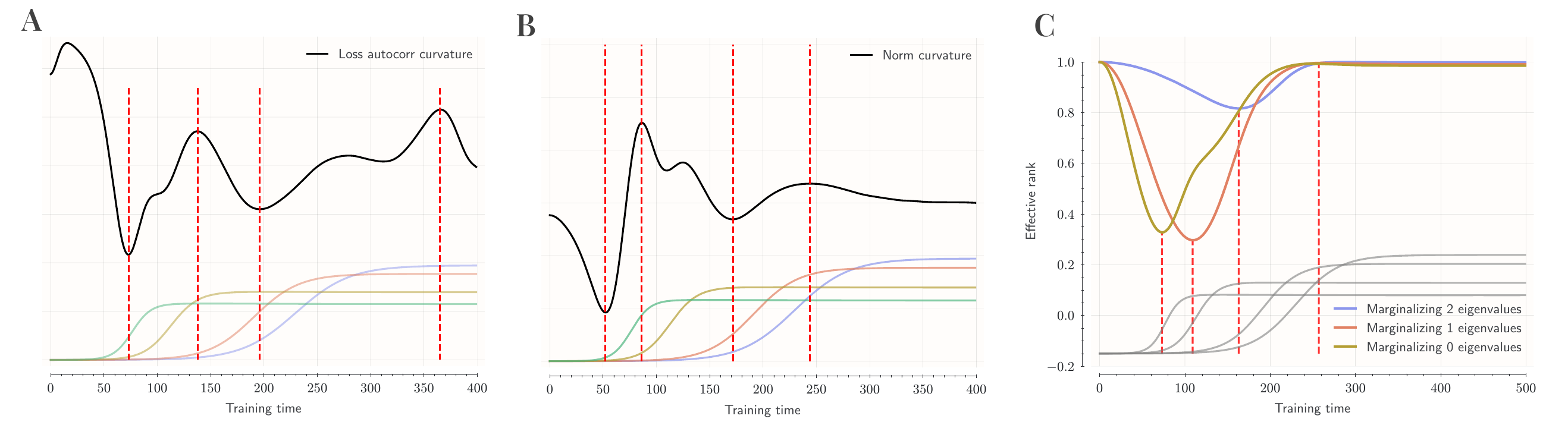} 
    \caption{Separation of timescale can be identified in (A) Curvature of loss autocorrelation function, (B) Curvature of parameter norm dynamics, and (C) Marginalized effective rank measure.}
    \label{fig:Figure4}
\end{figure*}

\section{Applications to Analysis of Non-linear Transformer Models}
\label{sec:applications}

While our analytical derivations rely on a simplified linear transformer, we hypothesize that key features of the learning dynamics—particularly the separation of timescales dictated by data statistics—may generalize to more complex, non-linear architectures. To investigate this, we extend insights from the solvable model to develop empirical probes for analyzing macroscopic features of learning dynamics in non-linear transformers.

\textbf{Probing Timescale Separation via Curvature.}
The solved loss dynamics (Eq.~\eqref{eq:loss_dynamics_main}) predict non-uniform learning rates across different spectral modes $\alpha$, leading to characteristic plateaus followed by cliffs. Detecting these transitions in the raw training loss of non-linear models is challenging due to the stochastic noise inherent in SGD, which obscures subtle inflection points. To mitigate this, we analyze the curvature of the loss autocorrelation function, defined as $A(\tau) = \mathbb{E}[L(t)L(t+\tau)]$. This metric acts as a smoothing filter, preserving macroscopic trend changes while suppressing high-frequency batch noise. Peaks in the second derivative of $A(\tau)$ indicate rapid changes in the rate of loss decay, corresponding to the "cliffs" where a new eigenmode is rapidly learned (Figure~\ref{fig:Figure4}A). Complementarily, we analyze the curvature of the Frobenius norm of the active parameter matrices ($\|p_2(t)\|_F$ or $\|q_1(t)\|_F$). Since the parameters $p_\alpha, q_\alpha$ evolve according to Eq.~\eqref{eq:decoupled_odes_main} with mode-dependent rates, their overall norm should also exhibit non-uniform changes. Peaks in the parameter norm curvature align well with those in the loss autocorrelation (Figure~\ref{fig:Figure4}B), reinforcing that macroscopic observables reflect the underlying timescale separation predicted by the theory.
\begin{figure*}[b!]
    \centering
    \includegraphics[width=\linewidth]{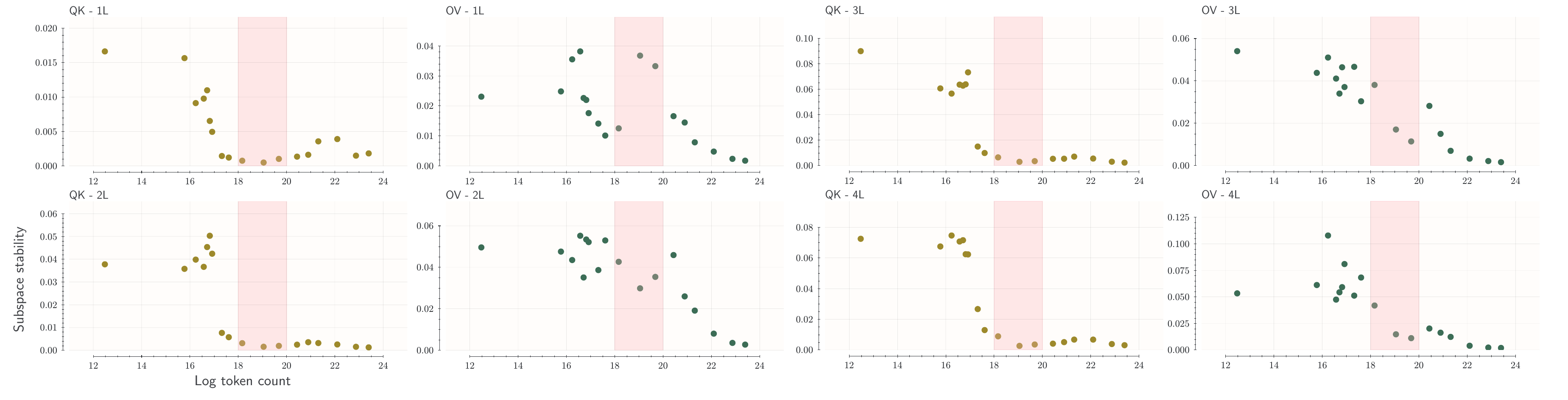} 
    \caption{Subspace Stabilization in Attention-Only Models (1-4 Layers). Spectral alignment metric (Subspace Distance, Eq.~\eqref{eq:subspace_distance_main}), averaged across OV and QK matrices, vs. training time. Lower values indicate subspace stability. The highlighted region marks ICL emergence. Subspace directions converge relatively early, preceding or coinciding with ICL emergence, especially in deeper models.}
    \label{fig:Figure6}
\end{figure*}

\textbf{Tracking Sequential Learning via Effective Rank.}
The exact solution $a_\alpha(t)$ (Eq.~\eqref{eq:a_alpha_solution_main}) describes the sequential "activation" or learning of different spectral modes. To track this process empirically in the weight matrices of non-linear models, we utilize the effective rank \citep{roy2007effective}, which measures the dimensionality or concentration of singular values ($s_i$) of a matrix $M$:
\begin{equation}
  \textrm{EffRank}(M) = \exp\left(-\sum_{i} \frac{s_i}{\sum_j s_j} \log\left(\frac{s_i}{\sum_j s_j}\right)\right).
  \label{eq:effective_rank_main}
\end{equation}
A lower effective rank implies energy concentration in fewer singular modes. Motivated by the theoretical prediction that dominant modes (large $s_\alpha$) learn first (small $\tau_\alpha$), we expect the effective rank to initially decrease as the model focuses on these modes, and subsequently increase as subdominant modes are learned. To better visualize this sequential process, we introduce "marginalized effective rank". This involves computing the effective rank repeatedly, each time excluding the singular value component corresponding to the mode assumed to have converged earliest. Dips in these marginalized curves should indicate the time points where specific spectral components undergo significant learning (Figure~\ref{fig:Figure4}C). As predicted by the sequential learning in Eq.~\eqref{eq:a_alpha_solution_main}, the effective rank initially dips, then recovers, and the marginalized curves reveal staggered learning across modes.
\begin{figure*}
    \centering
    \includegraphics[width=\linewidth]{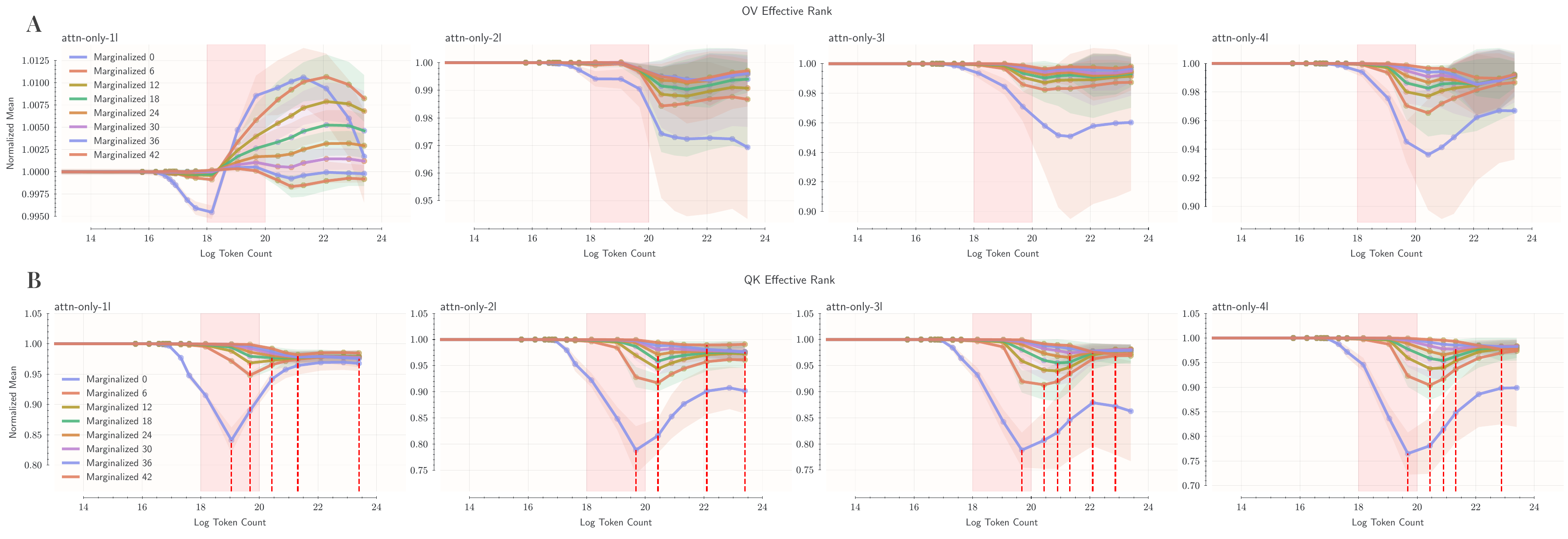} 
    \caption{Timescale Separation in Anthropic's Attention-Only Models (1-4 Layers). (A) Marginalized effective rank for Output-Value (OV) matrices. (B) Marginalized effective rank for Query-Key (QK) matrices. Averaged across layers/matrices. Highlighted region: ICL emergence. Note the characteristic dips preceding/coinciding with ICL emergence in deeper models (esp. QK), absent in the 1-layer model.}
    \label{fig:Figure5}
\end{figure*}
We applied these analyses to attention-only non-linear transformers (1-4 layers), similar to those exhibiting ICL capabilities \citep{olsson2022context}. Figure~\ref{fig:Figure5} displays the marginalized effective rank for Output-Value (OV) and Query-Key (QK) matrices. Consistent with the timescale separation hypothesis, models capable of ICL (2+ layers) show characteristic dips in the effective rank curves, particularly for QK matrices, around the time ICL performance emerges (highlighted region). The full set of marginalized curves reveals multiple minima occurring sequentially, supporting the idea derived from our linear model that learning proceeds in stages along different spectral dimensions. In contrast, the 1-layer model, which fails at this ICL task, shows a monotonic increase in OV effective rank, suggesting undifferentiated learning or noise fitting rather than structured, sequential learning. These empirical results suggest that the sequential, mode-dependent learning predicted by our solvable model qualitatively extends to these non-linear settings and correlates with the emergence of functional capabilities like ICL.

\textbf{Stability of Parameter Subspace.}
Our theoretical analysis separates the dynamics of parameter magnitudes (related to $a_\alpha = p_\alpha q_\alpha$) from the underlying parameter structure. The decoupled dynamics (Eq.~\eqref{eq:decoupled_odes_main}) allow for the possibility that the directions spanned by the singular vectors of weight matrices stabilize faster than the singular values reach their fixed points. To test this, we introduce a "Subspace Distance" metric, measuring the alignment between the subspace spanned by the singular vectors of a weight matrix $M(t)$ at time $t$ and its final state $M(\infty)$:
\begin{equation}
  \textrm{Sub. Dist.} \left[ M(t) \right] = \min_{A \in \mathbb{R}^{D \times D}} \left\| A M(t) - M(\infty) \right\|.
  \label{eq:subspace_distance_main}
\end{equation}
A smaller distance indicates subspace stabilization. Analyzing the attention-only models (Figure~\ref{fig:Figure6}), we find that the subspace distance (averaged across OV/QK matrices) decreases and stabilizes relatively early in training, often preceding or coinciding with the emergence of ICL and the primary dips in effective rank. This suggests that the network first identifies the relevant spectral directions and subsequently adjusts the magnitudes along these directions, consistent with a potential separation in learning dynamics for direction and magnitude.
\begin{figure*}
    \centering
    \includegraphics[width=0.8\linewidth]{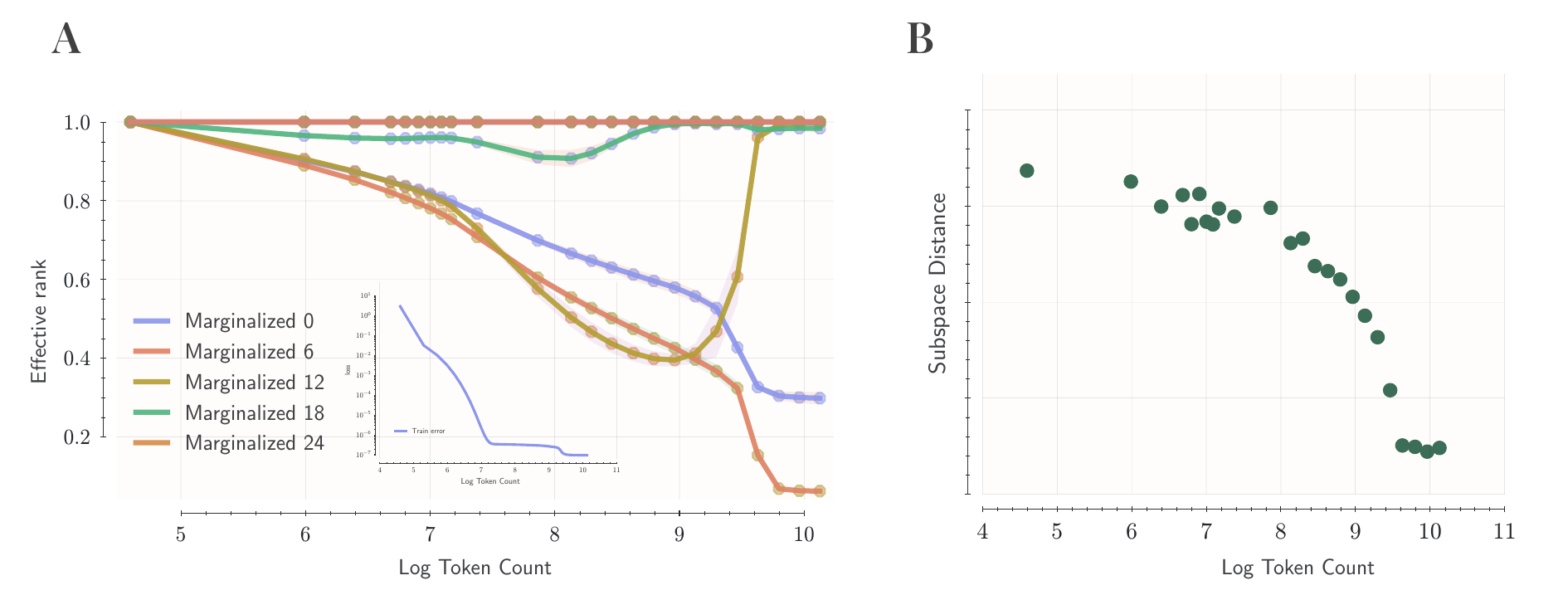} 
    \caption{Spectral Analysis of Grokking in Modular Arithmetic. (A) Delayed generalization characteristic of grokking (Train loss drops early, Test loss drops late). (B) Effective rank dynamics (OV matrix). (C) Subspace Distance (OV matrix). Note the late stabilization of the subspace, coinciding with the drop in test error (generalization), long after the training loss has converged.}
    \label{fig:Figure7}
\end{figure*}
This hypothesis is further tested by examining models exhibiting ``grokking'' in modular arithmetic tasks \citep{nanda2023grokking}, where generalization is significantly delayed relative to training performance saturation. Figure~\ref{fig:Figure3}B shows the characteristic delayed drop in test error. Analyzing the OV matrix dynamics, we observe that subspace stabilization, measured by the Subspace Distance (Figure~\ref{fig:Figure7}C), is markedly delayed. It occurs significantly after the training loss has minimized and coincides closely with the sharp improvement in generalization (test error drop). The effective rank (Figure~\ref{fig:Figure7}B) also shows dynamics linked to this delayed phase. This finding strongly suggests that grokking, in this context, is linked to a delayed convergence of the parameter subspace, providing a potential mechanistic explanation. Once the correct subspace is identified (albeit late), learning of magnitudes proceeds rapidly, leading to generalization.

In summary, empirical analyses of non-linear attention-based models using measures motivated by our solvable linear model reveal compelling qualitative parallels. Timescale separation, sequential mode-dependent learning, and the distinct role of subspace stabilization appear as potentially robust features extending beyond the linear regime. This suggests that insights from exactly solvable models can offer a valuable lens for interpreting complex learning dynamics in practical transformer architectures.

\section{Discussion}
\label{sec:discussion}

By analyzing an exactly solvable linear transformer model, we have derived the precise learning dynamics underlying its acquisition of in-context linear regression capabilities. Our central finding is that learning decomposes into independent dynamics along eigenmodes defined by the input data statistics. This decomposition naturally leads to a separation of timescales, where different modes are learned at rates inversely proportional to the square of the corresponding data eigenvalues ($\tau_\alpha \propto 1/s_\alpha^2$). This sequential learning manifests as distinct stages in parameter evolution and characteristic plateaus and cliffs in the loss function.

The learned computation at the fixed point (Eq.~\eqref{eq:learned_algorithm_main}) effectively implements the target linear map $W^\mu$ by applying an operator $\hat{\Sigma}_x [s^\infty(\Sigma_x)]^{-1}$ that acts to approximately whiten the input data $x_q$ relative to the context statistics $\hat{\Sigma}_x$. This operator effectively inverts the data covariance modulated by finite-context effects ($N$), allowing the model to isolate the underlying linear relationship. Broadly, our work demonstrates how analytical techniques from deep linear network theory \citep{saxe2014exact} can be extended to analyze context-dependent computations in transformers, revealing that timescale separation can arise directly from the interaction of attention parameters with data statistics, not just from network depth. Furthermore, the analysis revealed aspects of low-dimensional learning. First, parameter specialization occurs, where certain parameter blocks ($p_1, q_2$) remain "null" while others ($p_2, q_1$) actively drive learning. Second, a conservation law constrains the dynamics, reducing the effective dimensionality of the parameter trajectory and highlighting symmetries in the learning landscape.
Crucially, the qualitative features predicted by our solvable model—timescale separation, sequential mode learning, subspace stabilization—appear empirically relevant in non-linear attention-only models and grokking scenarios. The alignment between theoretical predictions and empirical observations using measures like effective rank and subspace distance suggests that the core mechanisms identified in the linear model offer valuable insights into the learning processes of more complex transformers.

Finally, this work suggests a principled approach to interpretability and evaluation. The theoretically grounded measures developed here—loss/norm curvature, marginalized effective rank, subspace distance—can empirically identify critical time points during training associated with significant learning events or potential shifts in model capabilities. Monitoring these metrics could provide a more informed strategy for timing behavioral evaluations, potentially capturing the emergence of new functionalities more effectively than relying solely on overall loss or accuracy measures. This offers a promising direction for bridging theoretical understanding with practical monitoring and analysis of large model training. Several avenues remain for future investigation. Incorporating non-linear attention mechanisms is essential to understand their impact on dynamics and potential deviations from the linear theory. Analyzing the effects of network depth and multiple attention heads could reveal how scaling influences ICL capabilities and learning complexity. Extending the analysis beyond linear regression to tasks involving non-linear functions or discrete structures is also a critical next step.


\section*{Acknowledgement}
We thank Jan Hendrick Kirchner for his guidance during the PIBBSS 2023 summer fellowship, and PIBBSS for continued support of the research work.

\bibliography{ICL}

\clearpage
\appendix

\renewcommand{\theequation}{A\arabic{equation}} 
\renewcommand{\thefigure}{A\arabic{figure}}
\renewcommand{\thetable}{A\arabic{table}}
\setcounter{equation}{0} 
\setcounter{figure}{0}
\setcounter{table}{0}

\section{Model Setup}
\label{sec:model_setup}

\subsection{Model Architecture}
\label{sec:model_architecture}
We consider a single-layer transformer network with a single linear attention head.  The attention mechanism employed is purely linear, omitting the softmax normalization typically found in standard attention mechanisms.

\subsection{Variables}
\label{sec:model_variables}
We define the key variables used in our model as follows:
\begin{itemize}
    \item \textbf{Input Data Pairs:} We are given $N$ pairs of $d$-dimensional data points $(x_i, y_i) \in \mathbb{R}^{d} \times \mathbb{R}^d$ for $i = 1, \dots, N$, and a query point $x_q \in \mathbb{R}^d$.
    \item \textbf{Embedding Dimension:} The embedding dimension is set to $D = 2d$.
    \item \textbf{Input Sequence Matrix:} For an input sequence of $N$ data pairs and a query point $(x_1, y_1, \dots, x_N, y_n, x_q)$, the input to the transformer is constructed as a matrix $Z \in \mathbb{R}^{D \times (N + 1)}$.  Specifically, $Z$ is defined as:
    \begin{equation}
        Z =
        \begin{bmatrix}
            x_1 & x_2 & \dots & x_N & x_q \\
            y_1 & y_2 & \dots & y_N & \mathbf{0}_d
        \end{bmatrix} \in \mathbb{R}^{D \times (N+1)},
        \label{eq:input_matrix_Z}
    \end{equation}
    where $\mathbf{0}_d \in \mathbb{R}^d$ is a zero vector of dimension $d$.
    \item \textbf{Weight Matrices:}
    \begin{itemize}
        \item \textbf{Combined Query-Key Weight Matrix:}  $W^Q \in \mathbb{R}^{D \times D}$.  We define a submatrix $q \in \mathbb{R}^{D \times d}$ comprising the first $d$ columns of $W^Q$.  This submatrix is further partitioned as $q = \begin{pmatrix} q_1 \\ q_2 \end{pmatrix}$, where $q_1, q_2 \in \mathbb{R}^{d \times d}$.
        \item \textbf{Output Projection Weight Matrix:} $W^P \in \mathbb{R}^{D \times D}$. We define a submatrix $p^T \in \mathbb{R}^{d \times D}$ comprising the bottom $d$ rows of $W^P$. This submatrix is further partitioned as $p^T = \begin{pmatrix} p_1 & p_2 \end{pmatrix}$, where $p_1, p_2 \in \mathbb{R}^{d \times d}$.
    \end{itemize}
\end{itemize}

\subsection{Linear Transformer}
\label{sec:linear_transformer}
The output of the linear attention head, before the residual connection, is given by:
\begin{equation}
    O = W^P \left( \frac{Z Z^\top}{N} \right) W^Q Z \in \mathbb{R}^{D \times (N+1)},
    \label{eq:linear_attention_output}
\end{equation}
where the attention operation is normalized by $N$.

The output of the single-layer linear transformer network, denoted as $f(Z) \in \mathbb{R}^{D \times (N+1)}$, is obtained by adding a residual connection:
\begin{equation}
    f(Z) = Z + O = Z + W^P \left( \frac{Z Z^\top}{N} \right) W^Q Z \in \mathbb{R}^{D \times (N+1)}.
    \label{eq:transformer_output}
\end{equation}

\subsection{Predictor}
\label{sec:predictor}
We are interested in the model's prediction for the query input $x_q$. This prediction is derived from the $(N+1)$-th column of the transformer network's output, $f(Z)$. Specifically, the prediction $\hat{y} \in \mathbb{R}^d$ is obtained from the bottom $d$ elements of the $(N+1)$-th column of $f(Z)$.

Due to the specific embedding structure defined in Section~\ref{sec:model_variables} and the output reading mechanism, the prediction can be simplified to:
\begin{equation}
    \hat{y} = \left( p^{\top} \hat{\Gamma} q \right) x_q \in \mathbb{R}^d,
    \label{eq:predictor_output}
\end{equation}
where $\hat{\Gamma} = \frac{Z Z^{\top}}{N} \in \mathbb{R}^{D \times D}$ is the empirical covariance matrix of the input $Z$.  Note that the matrix product $(p^{\top} \hat{\Gamma} q)$ results in a $d \times d$ matrix. The explicit derivation of this simplification is provided in Section~\ref{sec:simplified_predictor}.

\subsection{Loss Function}
\label{sec:loss_function}
To train the model, we employ the mean squared error (MSE) loss function, defined as:
\begin{equation}
    \mathcal{L} = \frac{1}{2} \| \hat{y} - y_q \|_2^2 = \frac{1}{2} \left\| \left( p^{\top} \hat{\Gamma} q \right) x_q - y_q \right\|_2^2.
    \label{eq:mse_loss}
\end{equation}
This loss function quantifies the discrepancy between the model's prediction $\hat{y}$ and the ground truth $y_q$.

\subsection{Goal}
\label{sec:goal}
The primary objective is to mathematically analyze this exactly solvable model of a linear transformer for in-context regression. Specifically, we aim to understand the learning dynamics and mechanisms through which the model predicts $y_q$ based on the provided in-context examples $\{(x_i, y_i)\}_{i=1}^N$.

\subsection{Key Assumptions}
\label{sec:key_assumptions}
Our analysis relies on the following key assumptions regarding the model architecture and setup:
\begin{itemize}
    \item \textbf{Single-layer, single-head linear transformer.} We consider a simplified transformer architecture with only one layer and a single attention head, both of which are linear.
    \item \textbf{No softmax in attention mechanism.} The linear attention mechanism omits the softmax normalization, resulting in a purely linear operation.
    \item \textbf{Specific embedding mechanism without positional tokens.} We employ a specific input embedding structure as defined in Section~\ref{sec:model_variables}, without the use of positional tokens.
    \item \textbf{Prediction focused on the $(\boldsymbol{N+1})$-th output token.} We focus our analysis on the prediction derived from the $(N+1)$-th output token, corresponding to the query input $x_q$.
\end{itemize}

\subsection{Task Setup}
\label{sec:task_setup}
We aim to train the linear transformer to perform in-context linear regression. To this end, we generate training data by sampling random instances of linear regression problems.

\subsubsection{x-data Generation}
\label{sec:x_data_generation}
During training, the input data $x$ is sampled from a multivariate Gaussian distribution with zero mean and a diagonal covariance matrix $\Sigma_{x} \in \mathbb{R}^{d \times d}$. Formally,
\begin{equation}
    x \sim \mathcal{N}\left( \mathbf{0}_d, \Sigma_x \right).
    \label{eq:x_data_distribution}
\end{equation}
Here, $\mathbf{0}_d \in \mathbb{R}^d$ is a zero-mean vector, and $\Sigma_x = \text{diag}(\sigma_{x,1}^2, \dots, \sigma_{x,d}^2)$ is a diagonal covariance matrix, where $\sigma_{x,i}^2 > 0$ are the variances along each dimension.

\subsubsection{Transformation Matrix $W$ Generation}
\label{sec:W_matrix_generation}
The transformation matrix $W \in \mathbb{R}^{d \times d}$ is generated to define the linear regression task. First, we sample $d$ eigenvalues $\lambda_1, \dots, \lambda_d$ from a standard multivariate Gaussian distribution $\mathcal{N}(\mathbf{0}_d, I_d)$. We then construct a diagonal matrix $\Lambda = \text{diag}(\lambda_1, \dots, \lambda_d) \in \mathbb{R}^{d \times d}$. For a fixed orthogonal matrix $V \in \mathbb{R}^{d \times d}$, the transformation matrix $W$ is defined as:
\begin{equation}
    W = V \Lambda V^{\top} \in \mathbb{R}^{d \times d}.
    \label{eq:W_matrix_definition}
\end{equation}
Note that, due to the zero mean of the eigenvalues, the expected value of $W$ is $\mathbb{E}[W] = 0$. Furthermore, $\mathbb{E}[W W^{\top}] = V \mathbb{E}[\Lambda^2] V^{\top} = V I_d V^{\top} = V V^{\top}$, since $\mathbb{E}[\Lambda^2] = I_d$ when eigenvalues are sampled from standard Gaussian. We consider $P$ independent regression tasks, each defined by a transformation matrix $W^{\mu}$ for $\mu = 1, \dots, P$. These matrices $\{W^{\mu}\}_{\mu=1}^P$ are sampled at the beginning of training and remain fixed throughout the training process.

\subsubsection{y-data Generation and Empirical Covariance}
\label{sec:y_data_generation}
For each regression task $\mu$ defined by $W^{\mu}$, we generate the corresponding $y$-data. First, we sample $x_1, \dots, x_N, x_q \in \mathbb{R}^d$ from the distribution defined in Section~\ref{sec:x_data_generation}. Then, for each $x_i$ ($i=1, \dots, N$), we generate $y_i$ using the linear transformation $y_i = W^{\mu} x_i$. For the query point $x_q$, the ground truth $y_q$ is similarly given by $y_q = W^{\mu} x_q$.

For a specific task $\mu$, the input sequence matrix $Z^{\mu}$ is constructed as:
\begin{equation}
    Z^{\mu} = \begin{bmatrix}
        x_1 & x_2 & \dots & x_N & x_q \\
        W^{\mu} x_1 & W^{\mu} x_2 & \dots & W^{\mu} x_N & \mathbf{0}_d
    \end{bmatrix} \in \mathbb{R}^{D \times (N+1)}.
    \label{eq:input_matrix_Z_mu}
\end{equation}
The model's prediction is based on the empirical covariance matrix of $Z^{\mu}$, denoted as $\hat{\Gamma}^{\mu} = \frac{1}{N} Z^{\mu} (Z^{\mu})^{\top} \in \mathbb{R}^{D \times D}$. To derive an explicit form for $\hat{\Gamma}^{\mu}$, we first compute $Z^{\mu} (Z^{\mu})^{\top}$:
\begin{align*}
    Z^{\mu} (Z^{\mu})^{\top} &= \begin{bmatrix}
        x_1 & \dots & x_N & x_q \\
        W^{\mu} x_1 & \dots & W^{\mu} x_N & \mathbf{0}_d
    \end{bmatrix} \begin{bmatrix}
        x_1^{\top} & (W^{\mu} x_1)^{\top} \\
        \vdots & \vdots \\
        x_N^{\top} & (W^{\mu} x_N)^{\top} \\
        x_q^{\top} & \mathbf{0}_d^{\top}
    \end{bmatrix} \\
    &= \begin{bmatrix}
        \sum_{i=1}^N x_i x_i^{\top} + x_q x_q^{\top} & \sum_{i=1}^N x_i (W^{\mu} x_i)^{\top} \\
        \sum_{i=1}^N W^{\mu} x_i x_i^{\top} & \sum_{i=1}^N W^{\mu} x_i (W^{\mu} x_i)^{\top}
    \end{bmatrix} \\
    &= \begin{bmatrix}
        \sum_{i=1}^N x_i x_i^{\top} + x_q x_q^{\top} & \sum_{i=1}^N x_i x_i^{\top} (W^{\mu})^{\top} \\
        \sum_{i=1}^N W^{\mu} x_i x_i^{\top} & \sum_{i=1}^N W^{\mu} x_i x_i^{\top} (W^{\mu})^{\top}
    \end{bmatrix}.
\end{align*}
Thus, the empirical covariance matrix $\hat{\Gamma}^{\mu}$ is given by:
\begin{equation}
    \hat{\Gamma}^{\mu} = \frac{1}{N} Z^{\mu} (Z^{\mu})^{\top} = \begin{bmatrix}
        \frac{1}{N} \sum_{i=1}^N x_i x_i^{\top} + \frac{1}{N} x_q x_q^{\top} & \frac{1}{N} \sum_{i=1}^N x_i x_i^{\top} (W^{\mu})^{\top} \\
        \frac{1}{N} \sum_{i=1}^N W^{\mu} x_i x_i^{\top} & \frac{1}{N} \sum_{i=1}^N W^{\mu} x_i x_i^{\top} (W^{\mu})^{\top}
    \end{bmatrix} \in \mathbb{R}^{D \times D}.
    \label{eq:empirical_covariance_Gamma_mu}
\end{equation}
Let $\hat{\Sigma}_x = \frac{1}{N} \sum_{i=1}^N x_i x_i^{\top}$ denote the empirical covariance of the training $x$-data $\{x_i\}_{i=1}^N$. Then, $\hat{\Gamma}^{\mu}$ can be expressed in block form as:
\begin{equation}
    \hat{\Gamma}^{\mu} = \begin{bmatrix}
        \hat{\Sigma}_x + \frac{1}{N} x_q x_q^{\top} & \hat{\Sigma}_x (W^{\mu})^{\top} \\
        W^{\mu} \hat{\Sigma}_x & W^{\mu} \hat{\Sigma}_x (W^{\mu})^{\top}
    \end{bmatrix}.
    \label{eq:empirical_covariance_Gamma_mu_block}
\end{equation}
In some subsequent derivations, for simplicity of notation, we might approximate $\hat{\Gamma}^{\mu}$ by ignoring the term $\frac{1}{N} x_q x_q^{\top}$ in the top-left block, leading to:
\begin{equation}
    \hat{\Gamma}^{\mu} \approx \begin{bmatrix}
        \hat{\Sigma}_x & \hat{\Sigma}_x (W^{\mu})^{\top} \\
        W^{\mu} \hat{\Sigma}_x & W^{\mu} \hat{\Sigma}_x (W^{\mu})^{\top}
    \end{bmatrix}.
    \label{eq:approx_empirical_covariance_Gamma_mu_block}
\end{equation}
This approximation is valid when $N$ is sufficiently large such that the contribution of the query point $x_q$ to the empirical covariance is negligible compared to the contribution of the in-context examples $\{x_i\}_{i=1}^N$.

\subsubsection{Training Setup}
\label{sec:training_setup}
We train the model using stochastic gradient descent (SGD) with a learning rate $\lambda$. Training proceeds in task-centric epochs. In each epoch, we iterate through the set of $P$ regression tasks $\{W^{\mu}\}_{\mu=1}^P$. For each task $W^{\mu}$, we sample a batch of in-context examples $\{(x_i, y_i)\}_{i=1}^N$ and a query point $x_q$, and compute the loss function $\mathcal{L}^{\mu}$ as defined in Section~\ref{sec:loss_function}. We then update the model parameters using the gradient of $\mathcal{L}^{\mu}$ with respect to the parameters, accumulated over the batch.  A training epoch is completed once all $P$ tasks have been presented to the model. Using batches for gradient updates helps to reduce noise in the training process.

\subsection{Simplified Predictor}
\label{sec:simplified_predictor}
In this section, we derive the simplified form of the predictor given in Equation~\eqref{eq:predictor_output}. Recall from Equation~\eqref{eq:transformer_output} that the output of the linear transformer is given by:
\begin{equation}
    f(Z) = Z + W^P \left( \frac{Z Z^\top}{N} \right) W^Q Z = Z + W^P \hat{\Gamma} W^Q Z.
    \label{eq:transformer_output_repeated}
\end{equation}
We are interested in the prediction $\hat{y}$, which is obtained from the bottom $d$ elements of the $(N+1)$-th column of $f(Z)$.

First, let's consider the $(N+1)$-th column of the input matrix $Z$, which is given by $\begin{pmatrix} x_q \\ \mathbf{0}_d \end{pmatrix} \in \mathbb{R}^D$.  Then, the $(N+1)$-th column of the matrix product $W^Q Z$ is obtained by multiplying $W^Q$ with this column vector:
\begin{equation}
    W^Q \begin{pmatrix} x_q \\ \mathbf{0}_d \end{pmatrix} = \begin{pmatrix} q_1 \\ q_2 \end{pmatrix} x_q \in \mathbb{R}^D,
    \label{eq:WQZ_column_Nplus1}
\end{equation}
where we have used the partition of $W^Q = [q | * ]$ and $q = \begin{pmatrix} q_1 \\ q_2 \end{pmatrix}$ as defined in Section~\ref{sec:model_variables}.

Next, we consider the term $\hat{\Gamma} W^Q Z = \left( \frac{Z Z^\top}{N} \right) W^Q Z$. The $(N+1)$-th column of this product is obtained by multiplying the matrix $\hat{\Gamma}$ with the $(N+1)$-th column of $W^Q Z$, which we just computed in Equation~\eqref{eq:WQZ_column_Nplus1}. Thus, the $(N+1)$-th column of $\hat{\Gamma} W^Q Z$ is:
\begin{equation}
    \hat{\Gamma} \left( W^Q \begin{pmatrix} x_q \\ \mathbf{0}_d \end{pmatrix} \right) = \hat{\Gamma} \begin{pmatrix} q_1 \\ q_2 \end{pmatrix} x_q \in \mathbb{R}^D.
    \label{eq:GammaWQZ_column_Nplus1}
\end{equation}

Now, we consider the term $W^P \hat{\Gamma} W^Q Z$. The $(N+1)$-th column of this term is obtained by multiplying $W^P$ with the vector in Equation~\eqref{eq:GammaWQZ_column_Nplus1}. Using the partition of $W^P = \begin{pmatrix} * \\ p^T \end{pmatrix}$ and $p^T = \begin{pmatrix} p_1 & p_2 \end{pmatrix}$ from Section~\ref{sec:model_variables}, we have:
\begin{equation}
    W^P \left( \hat{\Gamma} \begin{pmatrix} q_1 \\ q_2 \end{pmatrix} x_q \right) = \begin{pmatrix} * \\ p^T \end{pmatrix} \hat{\Gamma} \begin{pmatrix} q_1 \\ q_2 \end{pmatrix} x_q = \begin{pmatrix} * \\ p^T \hat{\Gamma} \begin{pmatrix} q_1 \\ q_2 \end{pmatrix} x_q \end{pmatrix} \in \mathbb{R}^D,
    \label{eq:WPGammaWQZ_column_Nplus1}
\end{equation}
where $*$ denotes the top $d$ rows of the resulting vector, which are not relevant for our prediction.

Finally, we look at the $(N+1)$-th column of $f(Z) = Z + W^P \hat{\Gamma} W^Q Z$. This column is the sum of the $(N+1)$-th column of $Z$ and the $(N+1)$-th column of $W^P \hat{\Gamma} W^Q Z$:
\begin{equation}
    \begin{pmatrix} x_q \\ \mathbf{0}_d \end{pmatrix} + W^P \hat{\Gamma} \begin{pmatrix} q_1 \\ q_2 \end{pmatrix} x_q = \begin{pmatrix} x_q \\ \mathbf{0}_d \end{pmatrix} + \begin{pmatrix} * \\ p^T \hat{\Gamma} \begin{pmatrix} q_1 \\ q_2 \end{pmatrix} x_q \end{pmatrix} = \begin{pmatrix} * \\ \mathbf{0}_d + p^T \hat{\Gamma} \begin{pmatrix} q_1 \\ q_2 \end{pmatrix} x_q \end{pmatrix} \in \mathbb{R}^D.
    \label{eq:fZ_column_Nplus1}
\end{equation}
The prediction $\hat{y}$ is the bottom $d$ elements of this resulting column. Thus, we have:
\begin{equation}
    \hat{y} = \mathbf{0}_d + p^T \hat{\Gamma} \begin{pmatrix} q_1 \\ q_2 \end{pmatrix} x_q = p^T \hat{\Gamma} \begin{pmatrix} q_1 \\ q_2 \end{pmatrix} x_q = \begin{pmatrix} p_1 & p_2 \end{pmatrix} \hat{\Gamma} \begin{pmatrix} q_1 \\ q_2 \end{pmatrix} x_q \in \mathbb{R}^d.
    \label{eq:predictor_output_derived}
\end{equation}
Since $q = \begin{pmatrix} q_1 \\ q_2 \end{pmatrix}$ and $p^T = \begin{pmatrix} p_1 & p_2 \end{pmatrix}$, we can rewrite the prediction as:
\begin{equation}
    \hat{y} = \left( p^T \hat{\Gamma} q \right) x_q = \left( \begin{pmatrix} p_1 & p_2 \end{pmatrix} \hat{\Gamma} \begin{pmatrix} q_1 \\ q_2 \end{pmatrix} \right) x_q \in \mathbb{R}^d.
    \label{eq:predictor_output_final}
\end{equation}
This matches the simplified predictor form given in Equation~\eqref{eq:predictor_output}.

For a specific regression task $\mu$, using the approximated empirical covariance matrix from Equation~\eqref{eq:approx_empirical_covariance_Gamma_mu_block}, $\hat{\Gamma}^{\mu} \approx \begin{bmatrix} \hat{\Sigma}_x & \hat{\Sigma}_x (W^{\mu})^{\top} \\ W^{\mu} \hat{\Sigma}_x & W^{\mu} \hat{\Sigma}_x (W^{\mu})^{\top} \end{bmatrix}$, the prediction can be further expanded as:
\begin{align*}
    \hat{y}_{\mu} &= \left( p^{T} \hat{\Gamma}^{\mu} q \right) x_q \\
    &= \begin{pmatrix} p_1 & p_2 \end{pmatrix} \begin{bmatrix} \hat{\Sigma}_x & \hat{\Sigma}_x (W^{\mu})^{\top} \\ W^{\mu} \hat{\Sigma}_x & W^{\mu} \hat{\Sigma}_x (W^{\mu})^{\top} \end{bmatrix} \begin{pmatrix} q_1 \\ q_2 \end{pmatrix} x_q \\
    &= \begin{pmatrix} p_1 & p_2 \end{pmatrix} \begin{pmatrix} \hat{\Sigma}_x q_1 + \hat{\Sigma}_x (W^{\mu})^{\top} q_2 \\ W^{\mu} \hat{\Sigma}_x q_1 + W^{\mu} \hat{\Sigma}_x (W^{\mu})^{\top} q_2 \end{pmatrix} x_q \\
    &= \left( p_1 (\hat{\Sigma}_x q_1 + \hat{\Sigma}_x (W^{\mu})^{\top} q_2) + p_2 (W^{\mu} \hat{\Sigma}_x q_1 + W^{\mu} \hat{\Sigma}_x (W^{\mu})^{\top} q_2) \right) x_q \\
    &= \left( p_1 \hat{\Sigma}_x q_1 + p_1 \hat{\Sigma}_x (W^{\mu})^{\top} q_2 + p_2 W^{\mu} \hat{\Sigma}_x q_1 + p_2 W^{\mu} \hat{\Sigma}_x (W^{\mu})^{\top} q_2 \right) x_q \\
    &= \left( (p_1 \hat{\Sigma}_x + p_2 W^{\mu} \hat{\Sigma}_x) q_1 + (p_1 \hat{\Sigma}_x (W^{\mu})^{\top} + p_2 W^{\mu} \hat{\Sigma}_x (W^{\mu})^{\top}) q_2 \right) x_q \\
    &= \left( (p_1 + p_2 W^{\mu}) \hat{\Sigma}_x q_1 + (p_1 + p_2 W^{\mu}) \hat{\Sigma}_x (W^{\mu})^{\top} q_2 \right) x_q \\
    &= \left( (p_1 + p_2 W^{\mu}) \hat{\Sigma}_x (q_1 + (W^{\mu})^{\top} q_2) \right) x_q.
    \label{eq:predictor_output_task_mu}
\end{align*}
Thus, for task $\mu$, the simplified predictor is given by:
\begin{equation}
    \hat{y}_{\mu} = \left( (p_1 + p_2 W^{\mu}) \hat{\Sigma}_x (q_1 + (W^{\mu})^{\top} q_2) \right) x_q \in \mathbb{R}^d.
    \label{eq:predictor_output_task_mu_final}
\end{equation}

When we initialize $p_1 = 0$ and $q_2 = 0$, the predictor for task $\mu$ further simplifies to:
\begin{equation}
    \hat{y}_{\mu} = \left( p_2 W^{\mu} \hat{\Sigma}_x q_1 \right) x_q \in \mathbb{R}^d.
    \label{eq:predictor_output_task_mu_initialization}
\end{equation}
This simplified form will be used in the subsequent analysis of the learning dynamics.

\subsection{Null Parameter Analysis}
\label{sec:null_parameters}
In this section, we demonstrate that with specific initial conditions, namely $p_1 = 0$ and $q_2 = 0$, the expected gradients with respect to $p_1$ and $q_2$ are zero at the beginning of training. This result justifies setting these parameters to zero throughout the training process, thereby simplifying the predictor.

Recall the loss function for a single task $\mu$ is given by:
\begin{equation}
    \mathcal{L}^{\mu} = \frac{1}{2} \| \hat{y}_{\mu} - y_q^{\mu} \|_2^2 = \frac{1}{2} \left\| \left( p^{T} \hat{\Gamma}^{\mu} q \right) x_q - y_q^{\mu} \right\|_2^2,
    \label{eq:loss_function_repeated}
\end{equation}
and the simplified predictor is:
\begin{equation}
    \hat{y}_{\mu} = \left( p^{T} \hat{\Gamma}^{\mu} q \right) x_q = \left( \begin{pmatrix} p_1 & p_2 \end{pmatrix} \hat{\Gamma}^{\mu} \begin{pmatrix} q_1 \\ q_2 \end{pmatrix} \right) x_q = \left( (p_1 + p_2 W^{\mu}) \hat{\Sigma}_x (q_1 + (W^{\mu})^{\top} q_2) \right) x_q,
    \label{eq:predictor_output_task_mu_repeated}
\end{equation}
where we have used the approximated empirical covariance matrix $\hat{\Gamma}^{\mu} \approx \begin{bmatrix} \hat{\Sigma}_x & \hat{\Sigma}_x (W^{\mu})^{\top} \\ W^{\mu} \hat{\Sigma}_x & W^{\mu} \hat{\Sigma}_x (W^{\mu})^{\top} \end{bmatrix}$ and expanded the matrix product. For simplicity in gradient derivation, let us define an intermediate term $A^{\mu} \in \mathbb{R}^d$ as:
\begin{equation}
    A^{\mu} = \hat{\Sigma}_x (q_1 + (W^{\mu})^{\top} q_2) x_q.
    \label{eq:A_mu_definition}
\end{equation}
Then the predictor can be written as:
\begin{equation}
    \hat{y}_{\mu} = (p_1 + p_2 W^{\mu}) A^{\mu}.
    \label{eq:predictor_output_task_mu_A}
\end{equation}
And the loss function becomes:
\begin{equation}
    \mathcal{L}^{\mu} = \frac{1}{2} \| (p_1 + p_2 W^{\mu}) A^{\mu} - W^{\mu} x_q \|_2^2.
    \label{eq:loss_function_task_mu_A}
\end{equation}

\subsubsection{Gradient with respect to $p_1$}
To compute the gradient of $\mathcal{L}^{\mu}$ with respect to $p_1$, we use the chain rule. Let $g(p_1) = (p_1 + p_2 W^{\mu}) A^{\mu} - W^{\mu} x_q$. Then $\mathcal{L}^{\mu} = \frac{1}{2} \| g(p_1) \|_2^2 = \frac{1}{2} g(p_1)^{\top} g(p_1)$. The gradient is given by:
\begin{equation}
    \frac{\partial \mathcal{L}^{\mu}}{\partial p_1} = \frac{\partial g(p_1)}{\partial p_1}^{\top} \frac{\partial \mathcal{L}^{\mu}}{\partial g(p_1)} = \left( \frac{\partial g(p_1)}{\partial p_1} \right)^{\top} g(p_1).
    \label{eq:gradient_chain_rule}
\end{equation}
We compute the derivative of $g(p_1)$ with respect to $p_1$. Since $p_1 \in \mathbb{R}^{d \times d}$, we consider the derivative with respect to $p_1$ as a linear transformation.
\begin{equation}
    \frac{\partial g(p_1)}{\partial p_1} = \frac{\partial}{\partial p_1} \left[ (p_1 + p_2 W^{\mu}) A^{\mu} - W^{\mu} x_q \right] = \frac{\partial}{\partial p_1} \left[ p_1 A^{\mu} + p_2 W^{\mu} A^{\mu} - W^{\mu} x_q \right] = A^{\mu \top}.
    \label{eq:dg_dp1}
\end{equation}
Substituting this and $g(p_1)$ into Equation~\eqref{eq:gradient_chain_rule}, we get:
\begin{equation}
    \frac{\partial \mathcal{L}^{\mu}}{\partial p_1} = (A^{\mu})^{\top} \left( (p_1 + p_2 W^{\mu}) A^{\mu} - W^{\mu} x_q \right).
    \label{eq:dL_dp1_intermediate}
\end{equation}
Substituting back the expression for $A^{\mu}$ from Equation~\eqref{eq:A_mu_definition}:
\begin{equation}
    \frac{\partial \mathcal{L}^{\mu}}{\partial p_1} = \left( \hat{\Sigma}_x (q_1 + (W^{\mu})^{\top} q_2) x_q \right)^{\top} \left( (p_1 + p_2 W^{\mu}) \hat{\Sigma}_x (q_1 + (W^{\mu})^{\top} q_2) x_q - W^{\mu} x_q \right).
    \label{eq:dL_dp1_full}
\end{equation}
The change in $p_1$ after one SGD update step, summing over all $P$ tasks, is:
\begin{equation}
    \Delta p_1 = - \lambda \sum_{\mu=1}^P \left( \hat{\Sigma}_x (q_1 + (W^{\mu})^{\top} q_2) x_q \right)^{\top} \left( (p_1 + p_2 W^{\mu}) \hat{\Sigma}_x (q_1 + (W^{\mu})^{\top} q_2) x_q - W^{\mu} x_q \right).
    \label{eq:Delta_p1_update}
\end{equation}
Now, we evaluate the expected update $\mathbb{E}_W[\Delta p_1]$ under the initial conditions $p_1 = 0$ and $q_2 = 0$. Substituting $p_1 = 0$ and $q_2 = 0$ into Equation~\eqref{eq:Delta_p1_update}, we get:
\begin{equation}
    \Delta p_1 = - \lambda \sum_{\mu=1}^P \left( \hat{\Sigma}_x (q_1 + (W^{\mu})^{\top} 0) x_q \right)^{\top} \left( (0 + p_2 W^{\mu}) \hat{\Sigma}_x (q_1 + (W^{\mu})^{\top} 0) x_q - W^{\mu} x_q \right).
    \label{eq:Delta_p1_initial_condition}
\end{equation}
Simplifying this expression:
\begin{align}
    \Delta p_1 &= - \lambda \sum_{\mu=1}^P \left( \hat{\Sigma}_x q_1 x_q \right)^{\top} \left( p_2 W^{\mu} \hat{\Sigma}_x q_1 x_q - W^{\mu} x_q \right) \\
    &= - \lambda \sum_{\mu=1}^P \left( x_q^{\top} q_1^{\top} \hat{\Sigma}_x^{\top} \right) \left( p_2 W^{\mu} \hat{\Sigma}_x q_1 x_q - W^{\mu} x_q \right) \\
    &= - \lambda \sum_{\mu=1}^P x_q^{\top} q_1^{\top} \hat{\Sigma}_x \left( p_2 W^{\mu} \hat{\Sigma}_x q_1 x_q - W^{\mu} x_q \right) \\
    &= - \lambda \sum_{\mu=1}^P \left( x_q^{\top} q_1^{\top} \hat{\Sigma}_x p_2 W^{\mu} \hat{\Sigma}_x q_1 x_q - x_q^{\top} q_1^{\top} \hat{\Sigma}_x W^{\mu} x_q \right) \\
    &= - \lambda \sum_{\mu=1}^P x_q^{\top} q_1^{\top} \hat{\Sigma}_x p_2 W^{\mu} \hat{\Sigma}_x q_1 x_q + \lambda \sum_{\mu=1}^P x_q^{\top} q_1^{\top} \hat{\Sigma}_x W^{\mu} x_q.
    \label{eq:Delta_p1_simplified}
\end{align}
Taking the expectation of $\Delta p_1$ with respect to $W$, and using the linearity of expectation:
\begin{equation}
    \mathbb{E}_W[\Delta p_1] = - \lambda \sum_{\mu=1}^P x_q^{\top} q_1^{\top} \hat{\Sigma}_x p_2 \mathbb{E}_W[W^{\mu}] \hat{\Sigma}_x q_1 x_q + \lambda \sum_{\mu=1}^P x_q^{\top} q_1^{\top} \hat{\Sigma}_x \mathbb{E}_W[W^{\mu}] x_q.
    \label{eq:E_Delta_p1}
\end{equation}
Since we have $\mathbb{E}_W[W^{\mu}] = 0$ from Section~\ref{sec:W_matrix_generation}, both terms in Equation~\eqref{eq:E_Delta_p1} become zero:
\begin{equation}
    \mathbb{E}_W[\Delta p_1] = 0.
    \label{eq:E_Delta_p1_zero}
\end{equation}
Thus, the expected gradient of $\mathcal{L}^{\mu}$ with respect to $p_1$ is zero when $p_1 = 0$ and $q_2 = 0$.

\subsubsection{Gradient with respect to $q_2$}
Now, we compute the gradient of $\mathcal{L}^{\mu}$ with respect to $q_2$. Using the chain rule again, we need to compute $\frac{\partial \hat{y}^{\mu}}{\partial q_2}$. From Equation~\eqref{eq:predictor_output_task_mu_A}, $\hat{y}^{\mu} = (p_1 + p_2 W^{\mu}) A^{\mu}$, and from Equation~\eqref{eq:A_mu_definition}, $A^{\mu} = \hat{\Sigma}_x (q_1 + (W^{\mu})^{\top} q_2) x_q$. Therefore,
\begin{align}
    \frac{\partial \hat{y}^{\mu}}{\partial q_2} &= \frac{\partial}{\partial q_2} \left[ (p_1 + p_2 W^{\mu}) \hat{\Sigma}_x (q_1 + (W^{\mu})^{\top} q_2) x_q \right] \\
    &= (p_1 + p_2 W^{\mu}) \hat{\Sigma}_x \frac{\partial}{\partial q_2} \left[ (q_1 + (W^{\mu})^{\top} q_2) x_q \right] \\
    &= (p_1 + p_2 W^{\mu}) \hat{\Sigma}_x \frac{\partial}{\partial q_2} \left[ q_1 x_q + (W^{\mu})^{\top} q_2 x_q \right] \\
    &= (p_1 + p_2 W^{\mu}) \hat{\Sigma}_x (W^{\mu})^{\top} x_q.
    \label{eq:dyhat_dq2}
\end{align}
The gradient of the loss $\mathcal{L}^{\mu}$ with respect to $q_2$ is then:
\begin{align}
    \frac{\partial \mathcal{L}^{\mu}}{\partial q_2} &= \left( \frac{\partial \hat{y}^{\mu}}{\partial q_2} \right)^{\top} (\hat{y}^{\mu} - y_q^{\mu}) \\
    &= \left( (p_1 + p_2 W^{\mu}) \hat{\Sigma}_x (W^{\mu})^{\top} x_q \right)^{\top} \left( (p_1 + p_2 W^{\mu}) \hat{\Sigma}_x (q_1 + (W^{\mu})^{\top} q_2) x_q - W^{\mu} x_q \right) \\
    &= x_q^{\top} (W^{\mu}) \hat{\Sigma}_x^{\top} (p_1 + p_2 W^{\mu})^{\top} \left( (p_1 + p_2 W^{\mu}) \hat{\Sigma}_x (q_1 + (W^{\mu})^{\top} q_2) x_q - W^{\mu} x_q \right) \\
    &= x_q^{\top} W^{\mu} \hat{\Sigma}_x (p_1^{\top} + (p_2 W^{\mu})^{\top}) \left( (p_1 + p_2 W^{\mu}) \hat{\Sigma}_x (q_1 + (W^{\mu})^{\top} q_2) x_q - W^{\mu} x_q \right),
    \label{eq:dL_dq2_full}
\end{align}
assuming $\hat{\Sigma}_x$ and $p_1, p_2, W^{\mu}$ are matrices such that transpose operation distributes linearly.  The change in $q_2$ after one SGD update step, summing over all $P$ tasks, is:
\begin{equation}
    \Delta q_2 = - \lambda \sum_{\mu=1}^P x_q^{\top} W^{\mu} \hat{\Sigma}_x (p_1^{\top} + (p_2 W^{\mu})^{\top}) \left( (p_1 + p_2 W^{\mu}) \hat{\Sigma}_x (q_1 + (W^{\mu})^{\top} q_2) x_q - W^{\mu} x_q \right).
    \label{eq:Delta_q2_update}
\end{equation}
Substituting $p_1 = 0$ and $q_2 = 0$ into Equation~\eqref{eq:Delta_q2_update}, we get:
\begin{equation}
    \Delta q_2 = - \lambda \sum_{\mu=1}^P x_q^{\top} W^{\mu} \hat{\Sigma}_x (0^{\top} + (p_2 W^{\mu})^{\top}) \left( (0 + p_2 W^{\mu}) \hat{\Sigma}_x (q_1 + (W^{\mu})^{\top} 0) x_q - W^{\mu} x_q \right).
    \label{eq:Delta_q2_initial_condition}
\end{equation}
Simplifying this expression:
\begin{align}
    \Delta q_2 &= - \lambda \sum_{\mu=1}^P x_q^{\top} W^{\mu} \hat{\Sigma}_x (p_2 W^{\mu})^{\top} \left( p_2 W^{\mu} \hat{\Sigma}_x q_1 x_q - W^{\mu} x_q \right) \\
    &= - \lambda \sum_{\mu=1}^P x_q^{\top} W^{\mu} \hat{\Sigma}_x (W^{\mu})^{\top} p_2^{\top} \left( p_2 W^{\mu} \hat{\Sigma}_x q_1 x_q - W^{\mu} x_q \right) \\
    &= - \lambda \sum_{\mu=1}^P \left( x_q^{\top} W^{\mu} \hat{\Sigma}_x (W^{\mu})^{\top} p_2^{\top} p_2 W^{\mu} \hat{\Sigma}_x q_1 x_q - x_q^{\top} W^{\mu} \hat{\Sigma}_x (W^{\mu})^{\top} p_2^{\top} W^{\mu} x_q \right) \\
    &= - \lambda \sum_{\mu=1}^P x_q^{\top} W^{\mu} \hat{\Sigma}_x (W^{\mu})^{\top} p_2^{\top} p_2 W^{\mu} \hat{\Sigma}_x q_1 x_q + \lambda \sum_{\mu=1}^P x_q^{\top} W^{\mu} \hat{\Sigma}_x (W^{\mu})^{\top} p_2^{\top} W^{\mu} x_q.
    \label{eq:Delta_q2_simplified}
\end{align}
Taking the expectation of $\Delta q_2$ with respect to $W$:
\begin{equation}
    \mathbb{E}_W[\Delta q_2] = - \lambda \sum_{\mu=1}^P x_q^{\top} \mathbb{E}_W[W^{\mu} \hat{\Sigma}_x (W^{\mu})^{\top} p_2^{\top} p_2 W^{\mu}] \hat{\Sigma}_x q_1 x_q + \lambda \sum_{\mu=1}^P x_q^{\top} \mathbb{E}_W[W^{\mu} \hat{\Sigma}_x (W^{\mu})^{\top} p_2^{\top} W^{\mu}] x_q.
    \label{eq:E_Delta_q2}
\end{equation}
Consider the term $\mathbb{E}_W[W^{\mu} \hat{\Sigma}_x (W^{\mu})^{\top} p_2^{\top} p_2 W^{\mu}]$. Since $W^{\mu}$ has zero mean and its elements are sampled from a Gaussian distribution, any term involving the expectation of an odd number of elements of $W^{\mu}$ will be zero.  The term $W^{\mu} \hat{\Sigma}_x (W^{\mu})^{\top} p_2^{\top} p_2 W^{\mu}$ involves a product of three instances of $W^{\mu}$.  While $\hat{\Sigma}_x$ and $p_2$ are constant with respect to the expectation over $W^{\mu}$, the expectation $\mathbb{E}_W[W^{\mu} \hat{\Sigma}_x (W^{\mu})^{\top} p_2^{\top} p_2 W^{\mu}]$ is a tensor of order 4, and each component of this tensor is an expectation of a product of three elements from $W^{\mu}$. For a centered Gaussian distribution, all odd moments are zero. Therefore, $\mathbb{E}_W[W^{\mu} \hat{\Sigma}_x (W^{\mu})^{\top} p_2^{\top} p_2 W^{\mu}] = 0$. Similarly, $\mathbb{E}_W[W^{\mu} \hat{\Sigma}_x (W^{\mu})^{\top} p_2^{\top} W^{\mu}] = 0$. Thus,
\begin{equation}
    \mathbb{E}_W[\Delta q_2] = 0.
    \label{eq:E_Delta_q2_zero}
\end{equation}
Hence, the expected gradient of $\mathcal{L}^{\mu}$ with respect to $q_2$ is also zero when $p_1 = 0$ and $q_2 = 0$.

\subsubsection{Simplified Predictor with $p_1 = 0$ and $q_2 = 0$}
Given that the expected gradients for $p_1$ and $q_2$ are zero at initialization and remain zero in expectation under SGD, we can fix $p_1 = 0$ and $q_2 = 0$ throughout the training. With these null parameters, the predictor from Equation~\eqref{eq:predictor_output_task_mu_final} simplifies to:
\begin{equation}
    \hat{y}_{\mu} = \left( (0 + p_2 W^{\mu}) \hat{\Sigma}_x (q_1 + (W^{\mu})^{\top} 0) \right) x_q = \left( p_2 W^{\mu} \hat{\Sigma}_x q_1 \right) x_q \in \mathbb{R}^d.
    \label{eq:predictor_output_task_mu_null_params}
\end{equation}
This simplified predictor, which depends only on the parameters $p_2$ and $q_1$, will be the focus of our subsequent analysis of the learning dynamics.

\subsection{Learning Dynamics of Parameters $p_2$ and $q_1$}
\label{sec:learning_parameters}
In this section, we derive the stochastic gradient descent (SGD) update rules for the parameters $p_2$ and $q_1$, and then transition to continuous-time dynamics to analyze the learning behavior. We will utilize the simplified predictor $\hat{y}_{\mu} = (p_2 W^{\mu} \hat{\Sigma}_x q_1) x_q$ and the loss function $\mathcal{L}^{\mu} = \frac{1}{2} \| p_2 W^{\mu} \hat{\Sigma}_x q_1 x_q - W^{\mu} x_q \|_2^2$, which are valid under the initial conditions $p_1 = 0$ and $q_2 = 0$ maintained throughout training as justified in Section~\ref{sec:null_parameters}.

\subsubsection{SGD Update for $p_2$}
The gradient of the loss function $\mathcal{L}^{\mu}$ with respect to $p_2$ for a given task $\mu$ is derived as follows. Let $h(p_2) = p_2 W^{\mu} \hat{\Sigma}_x q_1 x_q - W^{\mu} x_q$. Then $\mathcal{L}^{\mu} = \frac{1}{2} \| h(p_2) \|_2^2 = \frac{1}{2} h(p_2)^{\top} h(p_2)$. The gradient is given by:
\begin{equation}
    \nabla_{p_2} \mathcal{L}^{\mu} = \left( \frac{\partial h(p_2)}{\partial p_2} \right)^{\top} h(p_2).
    \label{eq:gradient_chain_rule_p2}
\end{equation}
We compute the derivative of $h(p_2)$ with respect to $p_2$:
\begin{equation}
    \frac{\partial h(p_2)}{\partial p_2} = \frac{\partial}{\partial p_2} \left[ p_2 W^{\mu} \hat{\Sigma}_x q_1 x_q - W^{\mu} x_q \right] = \frac{\partial}{\partial p_2} \left[ p_2 (W^{\mu} \hat{\Sigma}_x q_1 x_q) - W^{\mu} x_q \right] = (W^{\mu} \hat{\Sigma}_x q_1 x_q)^{\top} = x_q^{\top} q_1^{\top} \hat{\Sigma}_x^{\top} (W^{\mu})^{\top} = x_q^{\top} q_1^{\top} \hat{\Sigma}_x (W^{\mu})^{\top},
    \label{eq:dh_dp2}
\end{equation}
since $\hat{\Sigma}_x$ is symmetric. Substituting this into Equation~\eqref{eq:gradient_chain_rule_p2} and using $h(p_2) = p_2 W^{\mu} \hat{\Sigma}_x q_1 x_q - W^{\mu} x_q$, we obtain the gradient:
\begin{align}
    \nabla_{p_2} \mathcal{L}^{\mu} &= \left( x_q^{\top} q_1^{\top} \hat{\Sigma}_x (W^{\mu})^{\top} \right)^{\top} \left( p_2 W^{\mu} \hat{\Sigma}_x q_1 x_q - W^{\mu} x_q \right) \\
    &= (W^{\mu} \hat{\Sigma}_x q_1 x_q - W^{\mu} x_q) \left( x_q^{\top} q_1^{\top} \hat{\Sigma}_x (W^{\mu})^{\top} \right) \\
    &= \left( p_2 W^{\mu} \hat{\Sigma}_x q_1 x_q - W^{\mu} x_q \right) x_q^{\top} q_1^{\top} \hat{\Sigma}_x (W^{\mu})^{\top}.
    \label{eq:dL_dp2}
\end{align}
The SGD update rule for $p_2$ for a given task $\mu$ is thus:
\begin{align}
    \Delta p_2 (\mu) &= - \lambda \nabla_{p_{2}} \mathcal{L}_{\mu} \\
    &= - \lambda \left( p_2 W^{\mu} \hat{\Sigma}_x q_1 x_q - W^{\mu} x_q  \right) x_q^{\top} q_1^{\top} \hat{\Sigma}_x (W^{\mu})^{\top}.
    \label{eq:Delta_p2_mu}
\end{align}
Summing over all tasks $\mu = 1, \dots, P$ for one epoch update, we get:
\begin{align}
    \Delta p_2 &= \sum_{\mu=1}^P \Delta p_2 (\mu) \\
    &= - \lambda \sum_{\mu=1}^P \left( p_2 W^{\mu} \hat{\Sigma}_x q_1 x_q - W^{\mu} x_q  \right) x_q^{\top} q_1^{\top} \hat{\Sigma}_x (W^{\mu})^{\top} \\
    &= - \lambda \sum_{\mu=1}^P \left( p_2 W^{\mu} \hat{\Sigma}_x q_1 x_q x_q^{\top} q_1^{\top} \hat{\Sigma}_x (W^{\mu})^{\top} - W^{\mu} x_q x_q^{\top} q_1^{\top} \hat{\Sigma}_x (W^{\mu})^{\top} \right).
    \label{eq:Delta_p2_sum_tasks}
\end{align}

\subsubsection{Change of Variables and Assumptions for Tractable Dynamics}
To simplify the analysis and derive tractable learning dynamics, we introduce a change of variables and make further assumptions. We consider the singular value decomposition of the population covariance matrix $\Sigma_{x}$ and the empirical covariance matrix $\hat{\Sigma}_{x}$ of the input data $\{x_i\}_{i=1}^N$:
\begin{eqnarray}
    \Sigma_{x} &=& U S U^{\top}, \label{eq:Sigma_x_SVD} \\
    \hat{\Sigma}_{x} &=& U \hat{S} U^{\top}, \label{eq:Sigma_hat_x_SVD}
\end{eqnarray}
where $U \in \mathbb{R}^{d \times d}$ is an orthogonal matrix, $S = \text{diag}(s_1, \dots, s_d) \in \mathbb{R}^{d \times d}$ is a diagonal matrix of singular values of $\Sigma_x$, and $\hat{S} = \text{diag}(\hat{s}_1, \dots, \hat{s}_d) \in \mathbb{R}^{d \times d}$ is a diagonal matrix related to the empirical singular values. We further assume that the orthogonal matrix $V$ used in the generation of the transformation matrices $W^{\mu} = V \Lambda^{\mu} V^{\top}$ (Equation~\eqref{eq:W_matrix_definition}) is the same as $U$, i.e., $V = U$. Thus, we have $W^{\mu} = U \Lambda^{\mu} U^{\top}$.

We also assume that the parameters $p_2$ and $q_1$ are initialized and maintain the following form throughout training:
\begin{eqnarray}
    p_2 &=& U \bar{p}_2 U^{\top}, \label{eq:p2_decomposition} \\
    q_1 &=& U \bar{q}_1 U^{\top}, \label{eq:q1_decomposition}
\end{eqnarray}
where $\bar{p}_2$ and $\bar{q}_1$ are diagonal matrices.

Substituting these decompositions into the update rule for $\Delta p_2$ (Equation~\eqref{eq:Delta_p2_sum_tasks}), and noting that $(W^{\mu})^{\top} = (U \Lambda^{\mu} U^{\top})^{\top} = U (\Lambda^{\mu})^{\top} U^{\top} = U \Lambda^{\mu} U^{\top}$ since $\Lambda^{\mu}$ is a diagonal matrix, and similarly $\hat{\Sigma}_x^{\top} = \hat{\Sigma}_x$, we have:
\begin{align*}
    \Delta p_2 &= - \lambda \sum_{\mu=1}^P \left( (U \bar{p}_2 U^{\top}) (U \Lambda^{\mu} U^{\top}) (U \hat{S} U^{\top}) (U \bar{q}_1 U^{\top}) x_q x_q^{\top} (U \bar{q}_1 U^{\top})^{\top} (U \hat{S} U^{\top}) (U \Lambda^{\mu} U^{\top}) \right. \\
    & \qquad \qquad \left. - (U \Lambda^{\mu} U^{\top}) x_q x_q^{\top} (U \bar{q}_1 U^{\top})^{\top} (U \hat{S} U^{\top}) (U \Lambda^{\mu} U^{\top}) \right) \\
    &= - \lambda \sum_{\mu=1}^P \left( U \bar{p}_2 \Lambda^{\mu} \hat{S} \bar{q}_1 U^{\top} x_q x_q^{\top} U \bar{q}_1 \hat{S} \Lambda^{\mu} U^{\top} - U \Lambda^{\mu} U^{\top} x_q x_q^{\top} U \bar{q}_1 \hat{S} \Lambda^{\mu} U^{\top} \right),
\end{align*}
using the orthogonality of $U$, i.e., $U^{\top} U = U U^{\top} = I$. We know that the change in $p_2$ can also be decomposed as $\Delta p_2 = p_2(t+1) - p_2(t) = U \bar{p}_2(t+1) U^{\top} - U \bar{p}_2(t) U^{\top} = U (\bar{p}_2(t+1) - \bar{p}_2(t)) U^{\top} = U \Delta \bar{p}_2 U^{\top}$. Thus, $U \Delta \bar{p}_2 U^{\top}$ is equal to the expression above. Multiplying by $U^{\top}$ from the left and $U$ from the right, and using orthogonality of $U$, we get:
\begin{align*}
    \Delta \bar{p}_2 &= - \lambda \sum_{\mu=1}^P U^{\top} \left( U \bar{p}_2 \Lambda^{\mu} \hat{S} \bar{q}_1 U^{\top} x_q x_q^{\top} U \bar{q}_1 \hat{S} \Lambda^{\mu} U^{\top} - U \Lambda^{\mu} U^{\top} x_q x_q^{\top} U \bar{q}_1 \hat{S} \Lambda^{\mu} U^{\top} \right) U \\
    &= - \lambda \sum_{\mu=1}^P \left( \bar{p}_2 \Lambda^{\mu} \hat{S} \bar{q}_1 U^{\top} x_q x_q^{\top} U \bar{q}_1 \hat{S} \Lambda^{\mu} - \Lambda^{\mu} U^{\top} x_q x_q^{\top} U \bar{q}_1 \hat{S} \Lambda^{\mu} \right).
\end{align*}
Let $\bar{x}_q = U^{\top} x_q$. Then $x_q = U \bar{x}_q$ and $U^{\top} x_q x_q^{\top} U = U^{\top} (U \bar{x}_q) (U \bar{x}_q)^{\top} U = U^{\top} U \bar{x}_q \bar{x}_q^{\top} U^{\top} U = \bar{x}_q \bar{x}_q^{\top}$. Thus,
\begin{align}
    \Delta \bar{p}_2 &= - \lambda \sum_{\mu=1}^P \left( \bar{p}_2 \Lambda^{\mu} \hat{S} \bar{q}_1 \bar{x}_q \bar{x}_q^{\top} \bar{q}_1 \hat{S} \Lambda^{\mu} - \Lambda^{\mu} \bar{x}_q \bar{x}_q^{\top} \bar{q}_1 \hat{S} \Lambda^{\mu} \right) \\
    &= - \lambda \sum_{\mu=1}^P \left( \bar{p}_2 \Lambda^{\mu} \hat{S}^2 \bar{q}_1^2 \bar{x}_q \bar{x}_q^{\top} \Lambda^{\mu} - \Lambda^{\mu 2} \bar{q}_1 \bar{x}_q \bar{x}_q^{\top} \hat{S} \right),
    \label{eq:Delta_bar_p2_intermediate}
\end{align}
since $\hat{S}$ and $\bar{q}_1$ are diagonal matrices and commute.

Replacing the summation over tasks with expectation over $\Lambda$, we get the expected update for $\bar{p}_2$:
\begin{equation}
    \mathbb{E}_{\Lambda}[\Delta \bar{p}_2] = - \lambda P \mathbb{E}_{\Lambda} \left[ \bar{p}_2 \Lambda \hat{S}^2 \bar{q}_1^2 \bar{x}_q \bar{x}_q^{\top} \Lambda - \Lambda^2 \bar{q}_1 \bar{x}_q \bar{x}_q^{\top} \hat{S} \right].
    \label{eq:E_Delta_bar_p2_intermediate}
\end{equation}
Since all matrices in the expression are diagonal, they commute. Thus, we can rewrite Equation~\eqref{eq:E_Delta_bar_p2_intermediate} as:
\begin{equation}
    \mathbb{E}_{\Lambda}[\Delta \bar{p}_2] = - \lambda P \left( \bar{p}_2 \hat{S}^2 \bar{q}_1^2 \bar{x}_q \bar{x}_q^{\top} \mathbb{E}_{\Lambda}[\Lambda^2] - \mathbb{E}_{\Lambda}[\Lambda^2] \bar{q}_1 \bar{x}_q \bar{x}_q^{\top} \hat{S} \right).
    \label{eq:E_Delta_bar_p2_commute}
\end{equation}
Given that the eigenvalues in $\Lambda$ are sampled from a standard multivariate Gaussian distribution, $\mathbb{E}[\Lambda^2] = I$. Substituting this into Equation~\eqref{eq:E_Delta_bar_p2_commute}:
\begin{equation}
    \mathbb{E}_{\Lambda}[\Delta \bar{p}_2] = - \lambda P \left( \bar{p}_2 \hat{S}^2 \bar{q}_1^2 \bar{x}_q \bar{x}_q^{\top} - \bar{q}_1 \bar{x}_q \bar{x}_q^{\top} \hat{S} \right).
    \label{eq:E_Delta_bar_p2_Lambda_expected}
\end{equation}
Now, taking the expectation over $\bar{x}_q \bar{x}_q^{\top}$. Since $x_q \sim \mathcal{N}(0, \Sigma_x)$ and $\bar{x}_q = U^{\top} x_q$, we have $\bar{x}_q \sim \mathcal{N}(0, U^{\top} \Sigma_x U) = \mathcal{N}(0, S)$. The expectation of $\bar{x}_q \bar{x}_q^{\top}$ is $\mathbb{E}[\bar{x}_q \bar{x}_q^{\top}] = \text{Cov}(\bar{x}_q) = S$. Thus,
\begin{equation}
    \mathbb{E}_{\Lambda, x_q}[\Delta \bar{p}_2] = - \lambda P \left( \bar{p}_2 \hat{S}^2 \bar{q}_1^2 S - \bar{q}_1 S \hat{S} \right) = - \lambda P S \left( \bar{p}_2 \hat{S}^2 \bar{q}_1^2 - \bar{q}_1 \hat{S} \right).
    \label{eq:E_Delta_bar_p2_xq_expected}
\end{equation}
Finally, we take the expectation over the empirical covariance matrix $\hat{S}$. We know that $N \hat{\Sigma}_x = \sum_{i=1}^N x_i x_i^{\top}$. If $x_i \sim \mathcal{N}(0, \Sigma_x)$, then $N \hat{\Sigma}_x$ follows a Wishart distribution $W(N, \Sigma_x)$. In our diagonalized basis, $N \hat{S} = U^{\top} (N \hat{\Sigma}_x) U$ also behaves like a Wishart matrix related to $S$. For large $N$, $\hat{S} \approx S$. For expectation, we use the properties of Wishart distribution: $\mathbb{E}[\hat{S}] = S$ and $\mathbb{E}[\hat{S}^2] = \frac{N+1}{N}S^2 + \frac{\operatorname{Tr}(S)S}{N}$.
Thus, taking expectation over $\hat{S}$:
\begin{align}
    \mathbb{E}[\Delta \bar{p}_2] &= - \lambda P S \left( \bar{p}_2 \bar{q}_1^2 \mathbb{E}[\hat{S}^2] - \bar{q}_1 \mathbb{E}[\hat{S}] \right) \\
    &= - \lambda P S \left( \bar{p}_2 \bar{q}_1^2 \left( \frac{N+1}{N}S^2 + \frac{\operatorname{Tr}(S)S}{N} \right) - \bar{q}_1 S \right) \\
    &= - \lambda P S^2 \left( \bar{p}_2 \bar{q}_1^2 \left( \frac{N+1}{N}S + \frac{\operatorname{Tr}(S)}{N} I \right) - \bar{q}_1 I \right).
    \label{eq:E_Delta_bar_p2_Shat_expected}
\end{align}

Due to the symmetry between $p_2$ and $q_1$ in the predictor structure and loss function, the derivation for the expected update of $\bar{q}_1$, $\mathbb{E}[\Delta \bar{q}_1]$, will have a similar form with the roles of $\bar{p}_2$ and $\bar{q}_1$ interchanged.  Therefore, by symmetry, we can write:
\begin{equation}
    \mathbb{E}[\Delta \bar{q}_1] = - \lambda P S^2 \left( \bar{q}_1 \bar{p}_2^2 \left( \frac{N+1}{N}S + \frac{\operatorname{Tr}(S)}{N} I \right) - \bar{p}_2 I \right).
    \label{eq:E_Delta_bar_q1_Shat_expected}
\end{equation}

\subsubsection{Continuous-Time Dynamics}
To analyze the learning dynamics in continuous time, we replace the discrete updates $\mathbb{E}[\Delta \bar{p}_2]$ and $\mathbb{E}[\Delta \bar{q}_1]$ with their continuous-time counterparts $\frac{d \bar{p}_2}{dt}$ and $\frac{d \bar{q}_1}{dt}$, respectively.  This leads to the following system of ordinary differential equations describing the learning dynamics:
\begin{align}
    \frac{d \bar{p}_2}{dt} &= - \lambda P S^2 \left( \bar{p}_2 \bar{q}_1^2 \left( \frac{N+1}{N}S + \frac{\operatorname{Tr}(S)}{N} I \right) - \bar{q}_1 I \right), \label{eq:dp2_dt_continuous} \\
    \frac{d \bar{q}_1}{dt} &= - \lambda P S^2 \left( \bar{q}_1 \bar{p}_2^2 \left( \frac{N+1}{N}S + \frac{\operatorname{Tr}(S)}{N} I \right) - \bar{p}_2 I \right). \label{eq:dq1_dt_continuous}
\end{align}
Since $\bar{p}_2$, $\bar{q}_1$, and $S$ are diagonal matrices, these matrix equations decouple into $d$ independent systems of differential equations for the diagonal elements. Let $s_\alpha$ be the $\alpha$-th diagonal element of $S$, $p_\alpha$ be the $\alpha$-th diagonal element of $\bar{p}_2$, and $q_\alpha$ be the $\alpha$-th diagonal element of $\bar{q}_1$. Then, for each dimension $\alpha = 1, \dots, d$, we have the dynamical equations:
\begin{align}
    \frac{d p_\alpha}{dt} &= - \lambda P s_\alpha^2 \left( p_\alpha q_\alpha^2 \left( \frac{N+1}{N}s_\alpha + \frac{\operatorname{Tr}(S)}{N} \right) - q_\alpha \right), \label{eq:dp_alpha_dt_continuous} \\
    \frac{d q_\alpha}{dt} &= - \lambda P s_\alpha^2 \left( q_\alpha p_\alpha^2 \left( \frac{N+1}{N}s_\alpha + \frac{\operatorname{Tr}(S)}{N} \right) - p_\alpha \right). \label{eq:dq_alpha_dt_continuous}
\end{align}
These equations describe the continuous-time learning dynamics of the parameters $p_\alpha$ and $q_\alpha$ for each dimension $\alpha$.

\subsection{Dynamical Equations and Conserved Quantity}
\label{sec:dynamical_equations_conserved_quantity}
In this section, we analyze the continuous-time learning dynamics derived in Equations~\eqref{eq:dp2_dt_continuous} and \eqref{eq:dq1_dt_continuous} and identify a conserved quantity associated with these dynamics. We then simplify these equations further by considering the diagonal elements and solve for the learning trajectory.

\subsubsection{Conserved Quantity}
\label{sec:conserved_quantity}
An important observation regarding the learning dynamics is the existence of a conserved quantity, which remains constant throughout the training process. Let us define a quantity $\mathcal{C}$ as the difference of squared Frobenius norms of $\bar{p}_2$ and $\bar{q}_1$:
\begin{equation}
    \mathcal{C} = \| \bar{p}_2 \|_F^2 - \| \bar{q}_1 \|_F^2 = \text{Tr}(\bar{p}_2^{\top} \bar{p}_2) - \text{Tr}(\bar{q}_1^{\top} \bar{q}_1).
    \label{eq:conserved_quantity_C_matrix}
\end{equation}
Since $\bar{p}_2$ and $\bar{q}_1$ are diagonal matrices, $\| \bar{p}_2 \|_F^2 = \sum_{\alpha} p_{\alpha}^2$ and $\| \bar{q}_1 \|_F^2 = \sum_{\alpha} q_{\alpha}^2$, where $p_{\alpha}$ and $q_{\alpha}$ are the diagonal elements. Thus, $\mathcal{C} = \sum_{\alpha} (p_{\alpha}^2 - q_{\alpha}^2)$.

To show that $\mathcal{C}$ is conserved, we compute its time derivative $\frac{d\mathcal{C}}{dt}$ using the chain rule and the dynamical equations for $\frac{d \bar{p}_2}{dt}$ and $\frac{d \bar{q}_1}{dt}$ (Equations~\eqref{eq:dp2_dt_continuous} and \eqref{eq:dq1_dt_continuous}):
\begin{align*}
    \frac{d\mathcal{C}}{dt} &= \frac{d}{dt} \left( \| \bar{p}_2 \|_F^2 - \| \bar{q}_1 \|_F^2 \right) \\
    &= \frac{d}{dt} \text{Tr}(\bar{p}_2^{\top} \bar{p}_2) - \frac{d}{dt} \text{Tr}(\bar{q}_1^{\top} \bar{q}_1) \\
    &= 2 \text{Tr} \left( \bar{p}_2^{\top} \frac{d \bar{p}_2}{dt} \right) - 2 \text{Tr} \left( \bar{q}_1^{\top} \frac{d \bar{q}_1}{dt} \right) \\
    &= 2 \text{Tr} \left( \bar{p}_2 \frac{d \bar{p}_2}{dt} \right) - 2 \text{Tr} \left( \bar{q}_1 \frac{d \bar{q}_1}{dt} \right),
\end{align*}
since for diagonal matrices $\bar{p}_2^{\top} = \bar{p}_2$ and $\bar{q}_1^{\top} = \bar{q}_1$. Substituting the expressions for $\frac{d \bar{p}_2}{dt}$ and $\frac{d \bar{q}_1}{dt}$ from Equations~\eqref{eq:dp2_dt_continuous} and \eqref{eq:dq1_dt_continuous}:
\begin{align*}
    \frac{d\mathcal{C}}{dt} &= 2 \text{Tr} \left[ \bar{p}_2 \left( - \lambda P S^2 \left( \bar{p}_2 \bar{q}_1^2 \left( \frac{N+1}{N}S + \frac{\operatorname{Tr}(S)}{N} I \right) - \bar{q}_1 I \right) \right) \right] \\
    & \quad - 2 \text{Tr} \left[ \bar{q}_1 \left( - \lambda P S^2 \left( \bar{q}_1 \bar{p}_2^2 \left( \frac{N+1}{N}S + \frac{\operatorname{Tr}(S)}{N} I \right) - \bar{p}_2 I \right) \right) \right] \\
    &= -2 \lambda P \text{Tr} \left[ S^2 \bar{p}_2 \left( \bar{p}_2 \bar{q}_1^2 \left( \frac{N+1}{N}S + \frac{\operatorname{Tr}(S)}{N} I \right) - \bar{q}_1 I \right) \right] \\
    & \quad + 2 \lambda P \text{Tr} \left[ S^2 \bar{q}_1 \left( \bar{q}_1 \bar{p}_2^2 \left( \frac{N+1}{N}S + \frac{\operatorname{Tr}(S)}{N} I \right) - \bar{p}_2 I \right) \right] \\
    &= -2 \lambda P \text{Tr} \left[ S^2 \bar{p}_2^2 \bar{q}_1^2 \left( \frac{N+1}{N}S + \frac{\operatorname{Tr}(S)}{N} I \right) - S^2 \bar{p}_2 \bar{q}_1 I \right] \\
    & \quad + 2 \lambda P \text{Tr} \left[ S^2 \bar{q}_1^2 \bar{p}_2^2 \left( \frac{N+1}{N}S + \frac{\operatorname{Tr}(S)}{N} I \right) - S^2 \bar{q}_1 \bar{p}_2 I \right].
\end{align*}
Since matrix trace is linear and cyclic, and all matrices here are diagonal and thus commute, we have $\text{Tr}[S^2 \bar{p}_2^2 \bar{q}_1^2 (\cdot)] = \text{Tr}[S^2 \bar{q}_1^2 \bar{p}_2^2 (\cdot)]$ and $\text{Tr}[S^2 \bar{p}_2 \bar{q}_1 I] = \text{Tr}[S^2 \bar{q}_1 \bar{p}_2 I]$. Therefore, the terms cancel out exactly:
\begin{equation*}
    \frac{d\mathcal{C}}{dt} = 0.
\end{equation*}
This shows that $\mathcal{C} = \| \bar{p}_2 \|_F^2 - \| \bar{q}_1 \|_F^2$ is indeed a conserved quantity of the learning dynamics.

\subsubsection{Diagonal Element Dynamics and Solution}
\label{sec:diagonal_element_dynamics}
As noted earlier, the matrix dynamical equations decouple into $d$ independent systems for the diagonal elements. For each dimension $\alpha = 1, \dots, d$, the dynamical equations are given by Equations~\eqref{eq:dp_alpha_dt_continuous} and \eqref{eq:dq_alpha_dt_continuous}:
\begin{align}
    \frac{d p_\alpha}{dt} &= - \lambda P s_\alpha^2 \left( p_\alpha q_\alpha^2 \left( \frac{N+1}{N}s_\alpha + \frac{\operatorname{Tr}(S)}{N} \right) - q_\alpha \right), \label{eq:dp_alpha_dt_continuous_repeated} \\
    \frac{d q_\alpha}{dt} &= - \lambda P s_\alpha^2 \left( q_\alpha p_\alpha^2 \left( \frac{N+1}{N}s_\alpha + \frac{\operatorname{Tr}(S)}{N} \right) - p_\alpha \right). \label{eq:dq_alpha_dt_continuous_repeated}
\end{align}
Let us define a term $s_{\alpha}^{\infty} = \left( \frac{N+1}{N}s_\alpha + \frac{\operatorname{Tr}(S)}{N} \right)$. Then the dynamical equations can be rewritten as:
\begin{align}
    \frac{d p_\alpha}{dt} &= - \lambda P s_\alpha^2 \left( p_\alpha q_\alpha^2 s_{\alpha}^{\infty} - q_\alpha \right) = \lambda P s_\alpha^2 q_\alpha \left( 1 - p_\alpha q_\alpha s_{\alpha}^{\infty} \right), \label{eq:dp_alpha_dt_simplified} \\
    \frac{d q_\alpha}{dt} &= - \lambda P s_\alpha^2 \left( q_\alpha p_\alpha^2 s_{\alpha}^{\infty} - p_\alpha \right) = \lambda P s_\alpha^2 p_\alpha \left( 1 - p_\alpha q_\alpha s_{\alpha}^{\infty} \right). \label{eq:dq_alpha_dt_simplified}
\end{align}
We define an effective timescale for learning in each dimension $\alpha$ as $\tau_{\alpha} = \frac{1}{\lambda P s_{\alpha}^2}$. Then, the dynamics become:
\begin{align}
    \tau_{\alpha} \frac{d p_\alpha}{dt} &= q_\alpha \left( 1 - p_\alpha q_\alpha s_{\alpha}^{\infty} \right), \label{eq:dp_alpha_dt_timescale} \\
    \tau_{\alpha} \frac{d q_\alpha}{dt} &= p_\alpha \left( 1 - p_\alpha q_\alpha s_{\alpha}^{\infty} \right). \label{eq:dq_alpha_dt_timescale}
\end{align}
Observe that $\tau_{\alpha} p_\alpha \frac{d p_\alpha}{dt} = p_\alpha q_\alpha \left( 1 - p_\alpha q_\alpha s_{\alpha}^{\infty} \right)$ and $\tau_{\alpha} q_\alpha \frac{d q_\alpha}{dt} = p_\alpha q_\alpha \left( 1 - p_\alpha q_\alpha s_{\alpha}^{\infty} \right)$. Thus, $\tau_{\alpha} p_\alpha \frac{d p_\alpha}{dt} = \tau_{\alpha} q_\alpha \frac{d q_\alpha}{dt}$, which implies $p_\alpha \frac{d p_\alpha}{dt} = q_\alpha \frac{d q_\alpha}{dt}$, or $p_\alpha dp_\alpha = q_\alpha dq_\alpha$. Integrating both sides, we get $\int p_\alpha dp_\alpha = \int q_\alpha dq_\alpha$, leading to $\frac{1}{2} p_\alpha^2 = \frac{1}{2} q_\alpha^2 + C'_{\alpha}$, or $p_\alpha^2 - q_\alpha^2 = C_{\alpha}$, where $C_{\alpha} = 2C'_{\alpha}$ is a constant of integration. This is the element-wise conserved quantity, consistent with the matrix conserved quantity $\mathcal{C}$. If initialized with $p_{\alpha}(0) = q_{\alpha}(0)$, then $C_{\alpha} = 0$, and $p_{\alpha}^2(t) = q_{\alpha}^2(t)$ for all $t$. Assuming $p_{\alpha}(t)$ and $q_{\alpha}(t)$ maintain the same sign (if initialized non-negative, remain non-negative), we can consider $p_{\alpha}(t) = q_{\alpha}(t)$.

Let $a_{\alpha} = p_{\alpha} q_{\alpha} = p_{\alpha}^2 = q_{\alpha}^2$ (under the assumption $p_{\alpha}=q_{\alpha}$). Then, the dynamics for $a_{\alpha}$ can be derived as:
\begin{align*}
    \tau_{\alpha} \frac{d a_{\alpha}}{dt} &= \tau_{\alpha} \frac{d}{dt} (p_{\alpha} q_{\alpha}) = \tau_{\alpha} \left( \frac{d p_{\alpha}}{dt} q_{\alpha} + p_{\alpha} \frac{d q_{\alpha}}{dt} \right) \\
    &= q_{\alpha} \left( q_\alpha \left( 1 - p_\alpha q_\alpha s_{\alpha}^{\infty} \right) \right) + p_{\alpha} \left( p_\alpha \left( 1 - p_\alpha q_\alpha s_{\alpha}^{\infty} \right) \right) \\
    &= q_{\alpha}^2 \left( 1 - a_{\alpha} s_{\alpha}^{\infty} \right) + p_{\alpha}^2 \left( 1 - a_{\alpha} s_{\alpha}^{\infty} \right) \\
    &= (p_{\alpha}^2 + q_{\alpha}^2) \left( 1 - a_{\alpha} s_{\alpha}^{\infty} \right) = 2 a_{\alpha} \left( 1 - a_{\alpha} s_{\alpha}^{\infty} \right),
\end{align*}
under the assumption $p_\alpha = q_\alpha$, so $p_\alpha^2 = q_\alpha^2 = a_\alpha$.  The fixed point for $a_{\alpha}$ is obtained by setting $\frac{d a_{\alpha}}{dt} = 0$, which gives $2 a_{\alpha} \left( 1 - a_{\alpha} s_{\alpha}^{\infty} \right) = 0$. Non-trivial fixed point is $a_{\alpha}^{\infty} = \frac{1}{s_{\alpha}^{\infty}} = \frac{1}{\frac{N+1}{N}s_\alpha + \frac{\operatorname{Tr}(S)}{N}}$.

To solve the differential equation $\tau_{\alpha} \frac{d a_{\alpha}}{dt} = 2 a_{\alpha} \left( 1 - a_{\alpha} s_{\alpha}^{\infty} \right)$, we can separate variables:
\begin{equation*}
    \int \frac{d a_{\alpha}}{a_{\alpha} \left( 1 - a_{\alpha} s_{\alpha}^{\infty} \right)} = \int \frac{2}{\tau_{\alpha}} dt.
\end{equation*}
Using partial fraction decomposition, $\frac{1}{a_{\alpha} (1 - a_{\alpha} s_{\alpha}^{\infty})} = \frac{1}{a_{\alpha}} + \frac{s_{\alpha}^{\infty}}{1 - a_{\alpha} s_{\alpha}^{\infty}} = \frac{1}{a_{\alpha}} - \frac{-s_{\alpha}^{\infty}}{1 - a_{\alpha} s_{\alpha}^{\infty}}$. Thus,
\begin{align*}
    \int \left( \frac{1}{a_{\alpha}} + \frac{s_{\alpha}^{\infty}}{1 - a_{\alpha} s_{\alpha}^{\infty}} \right) d a_{\alpha} &= \frac{2t}{\tau_{\alpha}} + C''_{\alpha} \\
    \log |a_{\alpha}| - \log |1 - a_{\alpha} s_{\alpha}^{\infty}| &= \frac{2t}{\tau_{\alpha}} + C''_{\alpha} \\
    \log \left| \frac{a_{\alpha}}{1 - a_{\alpha} s_{\alpha}^{\infty}} \right| &= \frac{2t}{\tau_{\alpha}} + C''_{\alpha} \\
    \frac{a_{\alpha}}{1 - a_{\alpha} s_{\alpha}^{\infty}} &= e^{\frac{2t}{\tau_{\alpha}} + C''_{\alpha}} = e^{C''_{\alpha}} e^{\frac{2t}{\tau_{\alpha}}} = C'''_{\alpha} e^{\frac{2t}{\tau_{\alpha}}}.
\end{align*}
Let $a_{\alpha}(0) = a_{\alpha}^0$. Then $\frac{a_{\alpha}^0}{1 - a_{\alpha}^0 s_{\alpha}^{\infty}} = C'''_{\alpha}$. Thus, $\frac{a_{\alpha}}{1 - a_{\alpha} s_{\alpha}^{\infty}} = \frac{a_{\alpha}^0}{1 - a_{\alpha}^0 s_{\alpha}^{\infty}} e^{\frac{2t}{\tau_{\alpha}}}$.
Solving for $a_{\alpha}(t)$:
\begin{align*}
    a_{\alpha}(t) &= \frac{a_{\alpha}^0}{1 - a_{\alpha}^0 s_{\alpha}^{\infty}} e^{\frac{2t}{\tau_{\alpha}}} \left( 1 - a_{\alpha}(t) s_{\alpha}^{\infty} \right) \\
    a_{\alpha}(t) \left( 1 + \frac{a_{\alpha}^0}{1 - a_{\alpha}^0 s_{\alpha}^{\infty}} e^{\frac{2t}{\tau_{\alpha}}} s_{\alpha}^{\infty} \right) &= \frac{a_{\alpha}^0}{1 - a_{\alpha}^0 s_{\alpha}^{\infty}} e^{\frac{2t}{\tau_{\alpha}}} \\
    a_{\alpha}(t) &= \frac{\frac{a_{\alpha}^0}{1 - a_{\alpha}^0 s_{\alpha}^{\infty}} e^{\frac{2t}{\tau_{\alpha}}}}{1 + \frac{a_{\alpha}^0 s_{\alpha}^{\infty}}{1 - a_{\alpha}^0 s_{\alpha}^{\infty}} e^{\frac{2t}{\tau_{\alpha}}}} = \frac{a_{\alpha}^0 e^{\frac{2t}{\tau_{\alpha}}}}{(1 - a_{\alpha}^0 s_{\alpha}^{\infty}) + a_{\alpha}^0 s_{\alpha}^{\infty} e^{\frac{2t}{\tau_{\alpha}}}} \\
    &= \frac{a_{\alpha}^0}{ (1 - a_{\alpha}^0 s_{\alpha}^{\infty}) e^{-\frac{2t}{\tau_{\alpha}}} + a_{\alpha}^0 s_{\alpha}^{\infty} } = \frac{a_{\alpha}^0 / s_{\alpha}^{\infty}}{ ((1 - a_{\alpha}^0 s_{\alpha}^{\infty}) / s_{\alpha}^{\infty}) e^{-\frac{2t}{\tau_{\alpha}}} + a_{\alpha}^0 } \\
    &= \frac{a_{\alpha}^0 / s_{\alpha}^{\infty}}{ (\frac{1}{s_{\alpha}^{\infty}} - a_{\alpha}^0) e^{-\frac{2t}{\tau_{\alpha}}} + a_{\alpha}^0 } = \frac{a_{\alpha}^{\infty} a_{\alpha}^0}{ (a_{\alpha}^{\infty} - a_{\alpha}^0) e^{-\frac{2t}{\tau_{\alpha}}} + a_{\alpha}^0 } \\
    &= a_{\alpha}^{\infty} \frac{a_{\alpha}^0}{a_{\alpha}^0 + (a_{\alpha}^{\infty} - a_{\alpha}^0) e^{-\frac{2t}{\tau_{\alpha}}}}.
\end{align*}
Thus, the solution for $a_{\alpha}(t)$ is:
\begin{equation}
    a_{\alpha}(t) = a_{\alpha}^{\infty} \left( \frac{a_{\alpha}^0}{a_{\alpha}^0 + \left( a_{\alpha}^{\infty} - a_{\alpha}^0 \right) \exp \left( -\frac{2t}{\tau_{\alpha}} \right)} \right).
    \label{eq:solution_a_alpha_t}
\end{equation}
For small initial values $a_{\alpha}^0 \approx \epsilon$ and approaching the fixed point $a_{\alpha}(t) \approx a_{\alpha}^{\infty} (1 - \epsilon)$, the time $t$ required to reach close to the fixed point can be approximated by solving for $t$ in:
\begin{align*}
    \epsilon &\approx \left( a_{\alpha}^{\infty} - a_{\alpha}^0 \right) \exp \left( -\frac{2t}{\tau_{\alpha}} \right) \\
    \exp \left( \frac{2t}{\tau_{\alpha}} \right) &\approx \frac{a_{\alpha}^{\infty} - a_{\alpha}^0}{\epsilon} \approx \frac{a_{\alpha}^{\infty}}{\epsilon} = \frac{1}{s_{\alpha}^{\infty} \epsilon} \\
    \frac{2t}{\tau_{\alpha}} &\approx \log \left( \frac{1}{s_{\alpha}^{\infty} \epsilon} \right) \\
    t &\approx \frac{\tau_{\alpha}}{2} \log \left( \frac{1}{s_{\alpha}^{\infty} \epsilon} \right) = \frac{1}{2\lambda P s_{\alpha}^2} \log \left( \frac{1}{s_{\alpha}^{\infty} \epsilon} \right).
    \label{eq:time_to_fixed_point}
\end{align*}
This gives an estimate of the convergence time for each dimension $\alpha$.

\subsection{Loss Analysis}
\label{sec:loss_analysis}
In this section, we analyze the expected loss function of the linear transformer model. We first express the average loss per epoch and then simplify it using the change of variables and assumptions introduced in Section~\ref{sec:learning_parameters}. We will derive the expected loss in terms of the diagonal parameters $a_\alpha(t)$ and analyze the loss behavior as training progresses.

\subsubsection{Expected Loss Formulation}
The loss function for a single task $\mu$ is given by:
\begin{equation}
    \mathcal{L}^{\mu} = \frac{1}{2} \| \hat{y}_{\mu} - y_q^{\mu} \|_2^2 = \frac{1}{2} \| p_2 W^{\mu} \hat{\Sigma}_x q_1 x_q - W^{\mu} x_q \|_2^2.
    \label{eq:loss_function_task_mu_repeated_loss_analysis}
\end{equation}
The average loss per epoch over all $P$ tasks is:
\begin{equation}
    \mathcal{L} = \frac{1}{P} \sum_{\mu=1}^P \mathcal{L}^{\mu} = \frac{1}{2P} \sum_{\mu=1}^P \| p_2 W^{\mu} \hat{\Sigma}_x q_1 x_q - W^{\mu} x_q \|_2^2.
    \label{eq:average_loss_sum_tasks}
\end{equation}
Replacing the summation with expectation over the distribution of $W$ and $x_q$, we obtain the expected loss:
\begin{equation}
    \mathcal{L} = \frac{1}{2} \mathbb{E}_{W, x_q} \left[ \| p_2 W \hat{\Sigma}_x q_1 x_q - W x_q \|_2^2 \right].
    \label{eq:expected_loss_W_xq}
\end{equation}
Using the change of variables $p_2 = U \bar{p}_2 U^{\top}$, $q_1 = U \bar{q}_1 U^{\top}$, $W = U \Lambda U^{\top}$, $\hat{\Sigma}_x = U \hat{S} U^{\top}$, and $x_q = U \bar{x}_q$, the expected loss becomes:
\begin{align}
    \mathcal{L} &= \frac{1}{2} \mathbb{E}_{\Lambda, \bar{x}_q} \left[ \| (U \bar{p}_2 U^{\top}) (U \Lambda U^{\top}) (U \hat{S} U^{\top}) (U \bar{q}_1 U^{\top}) (U \bar{x}_q) - (U \Lambda U^{\top}) (U \bar{x}_q) \|_2^2 \right] \\
    &= \frac{1}{2} \mathbb{E}_{\Lambda, \bar{x}_q} \left[ \| U \bar{p}_2 \Lambda \hat{S} \bar{q}_1 \bar{x}_q - U \Lambda \bar{x}_q \|_2^2 \right] \\
    &= \frac{1}{2} \mathbb{E}_{\Lambda, \bar{x}_q} \left[ \| U (\bar{p}_2 \Lambda \hat{S} \bar{q}_1 \bar{x}_q - \Lambda \bar{x}_q) \|_2^2 \right].
    \label{eq:expected_loss_bar_vars_U}
\end{align}
Since the Euclidean norm is invariant under orthogonal transformations, $\| U v \|_2 = \| v \|_2$ for any vector $v$ and orthogonal matrix $U$. Thus, Equation~\eqref{eq:expected_loss_bar_vars_U} simplifies to:
\begin{equation}
    \mathcal{L} = \frac{1}{2} \mathbb{E}_{\Lambda, \bar{x}_q} \left[ \| \bar{p}_2 \Lambda \hat{S} \bar{q}_1 \bar{x}_q - \Lambda \bar{x}_q \|_2^2 \right].
    \label{eq:expected_loss_bar_vars}
\end{equation}
Under the assumption that $p_\alpha(t) = q_\alpha(t)$, we have $\bar{p}_2 = \bar{q}_1 = \bar{a}^{1/2}$, where $\bar{a} = \text{diag}(a_1, \dots, a_d)$. Then $\bar{p}_2 \bar{q}_1 = \bar{a}$. The expected loss becomes:
\begin{equation}
    \mathcal{L} = \frac{1}{2} \mathbb{E}_{\Lambda, \bar{x}_q} \left[ \| \bar{a} \Lambda \hat{S} \bar{x}_q - \Lambda \bar{x}_q \|_2^2 \right].
    \label{eq:expected_loss_bar_a}
\end{equation}

\subsubsection{Expansion and Term-wise Expectation}
Expanding the squared norm in Equation~\eqref{eq:expected_loss_bar_a}, we get:
\begin{align}
    \mathcal{L} &= \frac{1}{2} \mathbb{E}_{\Lambda, \bar{x}_q} \left[ (\bar{a} \Lambda \hat{S} \bar{x}_q - \Lambda \bar{x}_q)^{\top} (\bar{a} \Lambda \hat{S} \bar{x}_q - \Lambda \bar{x}_q) \right] \\
    &= \frac{1}{2} \mathbb{E}_{\Lambda, \bar{x}_q} \left[ (\bar{x}_q^{\top} \hat{S} \Lambda \bar{a} - \bar{x}_q^{\top} \Lambda) (\bar{a} \Lambda \hat{S} \bar{x}_q - \Lambda \bar{x}_q) \right] \\
    &= \frac{1}{2} \mathbb{E}_{\Lambda, \bar{x}_q} \left[ \bar{x}_q^{\top} \hat{S} \Lambda \bar{a}^2 \Lambda \hat{S} \bar{x}_q - \bar{x}_q^{\top} \hat{S} \Lambda \bar{a} \Lambda \bar{x}_q - \bar{x}_q^{\top} \Lambda \bar{a} \Lambda \hat{S} \bar{x}_q + \bar{x}_q^{\top} \Lambda^2 \bar{x}_q \right] \\
    &= \frac{1}{2} \mathbb{E}_{\Lambda, \bar{x}_q} \left[ \bar{x}_q^{\top} \hat{S} \Lambda \bar{a}^2 \Lambda \hat{S} \bar{x}_q - 2 \bar{x}_q^{\top} \hat{S} \Lambda \bar{a} \Lambda \bar{x}_q + \bar{x}_q^{\top} \Lambda^2 \bar{x}_q \right],
    \label{eq:expected_loss_expanded}
\end{align}
where we used the fact that $\bar{x}_q^{\top} \hat{S} \Lambda \bar{a} \Lambda \bar{x}_q = (\bar{x}_q^{\top} \hat{S} \Lambda \bar{a} \Lambda \bar{x}_q)^{\top} = \bar{x}_q^{\top} \Lambda^{\top} \bar{a}^{\top} \Lambda^{\top} \hat{S}^{\top} \bar{x}_q = \bar{x}_q^{\top} \Lambda \bar{a} \Lambda \hat{S} \bar{x}_q$ since all matrices are diagonal and symmetric. We can analyze each term separately using trace operator and properties of expectation.
Let $E_1 = \mathbb{E}[\bar{x}_q^{\top} \hat{S} \Lambda \bar{a}^2 \Lambda \hat{S} \bar{x}_q]$, $E_2 = \mathbb{E}[\bar{x}_q^{\top} \hat{S} \Lambda \bar{a} \Lambda \bar{x}_q]$, and $E_3 = \mathbb{E}[\bar{x}_q^{\top} \Lambda^2 \bar{x}_q]$. Then $\mathcal{L} = \frac{1}{2} (E_1 - 2 E_2 + E_3)$.

For $E_1$:
\begin{align*}
    E_1 &= \mathbb{E}_{\Lambda, \bar{x}_q} [\bar{x}_q^{\top} \hat{S} \Lambda \bar{a}^2 \Lambda \hat{S} \bar{x}_q] = \mathbb{E}_{\Lambda, \bar{x}_q} [\text{Tr}(\bar{x}_q^{\top} \hat{S} \Lambda \bar{a}^2 \Lambda \hat{S} \bar{x}_q)] = \mathbb{E}_{\Lambda, \bar{x}_q} [\text{Tr}(\hat{S} \Lambda \bar{a}^2 \Lambda \hat{S} \bar{x}_q \bar{x}_q^{\top})] \\
    &= \mathbb{E}_{\Lambda} [\text{Tr}(\hat{S} \Lambda \bar{a}^2 \Lambda \hat{S} \mathbb{E}_{\bar{x}_q | \Lambda} [\bar{x}_q \bar{x}_q^{\top}])] = \mathbb{E}_{\Lambda} [\text{Tr}(\hat{S} \Lambda \bar{a}^2 \Lambda \hat{S} S)],
\end{align*}
since $\mathbb{E}_{\bar{x}_q}[\bar{x}_q \bar{x}_q^{\top}] = S$ and is independent of $\Lambda$. As trace is linear and cyclic, and matrices are diagonal,
\begin{align*}
    E_1 &= \text{Tr}(\mathbb{E}_{\Lambda} [\hat{S} \Lambda \bar{a}^2 \Lambda \hat{S}] S) = \text{Tr}(\mathbb{E}_{\Lambda} [\Lambda^2] \hat{S}^2 \bar{a}^2 S) = \text{Tr}(I \mathbb{E}[\hat{S}^2] \bar{a}^2 S) = \text{Tr}(\mathbb{E}[\hat{S}^2] \bar{a}^2 S),
\end{align*}
using $\mathbb{E}[\Lambda^2] = I$ and commutativity. Using $\mathbb{E}[\hat{S}^2] = \frac{N+1}{N}S^2 + \frac{\operatorname{Tr}(S)}{N}S$,
\begin{equation}
    E_1 = \text{Tr} \left( \left( \frac{N+1}{N}S^2 + \frac{\operatorname{Tr}(S)}{N}S \right) \bar{a}^2 S \right).
    \label{eq:E1_result}
\end{equation}

For $E_2$:
\begin{align*}
    E_2 &= \mathbb{E}_{\Lambda, \bar{x}_q} [\bar{x}_q^{\top} \hat{S} \Lambda \bar{a} \Lambda \bar{x}_q] = \mathbb{E}_{\Lambda, \bar{x}_q} [\text{Tr}(\bar{x}_q^{\top} \hat{S} \Lambda \bar{a} \Lambda \bar{x}_q)] = \mathbb{E}_{\Lambda, \bar{x}_q} [\text{Tr}(\hat{S} \Lambda \bar{a} \Lambda \bar{x}_q \bar{x}_q^{\top})] \\
    &= \mathbb{E}_{\Lambda} [\text{Tr}(\hat{S} \Lambda \bar{a} \Lambda \mathbb{E}_{\bar{x}_q | \Lambda} [\bar{x}_q \bar{x}_q^{\top}])] = \mathbb{E}_{\Lambda} [\text{Tr}(\hat{S} \Lambda \bar{a} \Lambda S)] = \text{Tr}(\mathbb{E}_{\Lambda} [\hat{S} \Lambda \bar{a} \Lambda] S) \\
    &= \text{Tr}(\mathbb{E}_{\Lambda} [\Lambda^2] \hat{S} \bar{a} S) = \text{Tr}(I \mathbb{E}[\hat{S}] \bar{a} S) = \text{Tr}(\mathbb{E}[\hat{S}] \bar{a} S) = \text{Tr}(S \bar{a} S) = \text{Tr}(\bar{a} S^2),
\end{align*}
using $\mathbb{E}[\Lambda^2] = I$ and $\mathbb{E}[\hat{S}] = S$.
\begin{equation}
    E_2 = \text{Tr}(\bar{a} S^2).
    \label{eq:E2_result}
\end{equation}

For $E_3$:
\begin{align*}
    E_3 &= \mathbb{E}_{\Lambda, \bar{x}_q} [\bar{x}_q^{\top} \Lambda^2 \bar{x}_q] = \mathbb{E}_{\Lambda, \bar{x}_q} [\text{Tr}(\bar{x}_q^{\top} \Lambda^2 \bar{x}_q)] = \mathbb{E}_{\Lambda, \bar{x}_q} [\text{Tr}(\Lambda^2 \bar{x}_q \bar{x}_q^{\top})] = \text{Tr}(\mathbb{E}_{\Lambda, \bar{x}_q} [\Lambda^2 \bar{x}_q \bar{x}_q^{\top}]) \\
    &= \text{Tr}(\mathbb{E}_{\Lambda} [\Lambda^2] \mathbb{E}_{\bar{x}_q} [\bar{x}_q \bar{x}_q^{\top}]) = \text{Tr}(I \cdot S) = \text{Tr}(S).
\end{align*}
\begin{equation}
    E_3 = \text{Tr}(S).
    \label{eq:E3_result}
\end{equation}

Substituting Equations~\eqref{eq:E1_result}, \eqref{eq:E2_result}, and \eqref{eq:E3_result} into $\mathcal{L} = \frac{1}{2} (E_1 - 2 E_2 + E_3)$:
\begin{align}
    \mathcal{L} &= \frac{1}{2} \left( \text{Tr} \left( \left( \frac{N+1}{N} S^2 + \frac{\operatorname{Tr}(S)}{N} S \right) \bar{a}^2 S \right) - 2 \text{Tr} \left( \bar{a} S^2 \right) + \text{Tr}(S) \right) \\
    &= \frac{1}{2} \sum_\alpha \left( \left( \frac{N+1}{N} s_\alpha^2 + \frac{\operatorname{Tr}(S)}{N} s_\alpha \right) a_\alpha^2 s_\alpha - 2 a_\alpha s_\alpha^2 + s_\alpha \right) \\
    &= \frac{1}{2} \sum_\alpha s_\alpha \left( a_\alpha^2 s_\alpha \left( \frac{N+1}{N} s_\alpha + \frac{\operatorname{Tr}(S)}{N} \right) - 2 a_\alpha s_\alpha + 1 \right).
    \label{eq:expected_loss_final_a_alpha}
\end{align}
Using $a_\alpha^\infty = \frac{1}{\frac{N+1}{N} s_\alpha + \frac{\operatorname{Tr}(S)}{N}}$, or $\frac{1}{a_\alpha^\infty} = \frac{N+1}{N} s_\alpha + \frac{\operatorname{Tr}(S)}{N}$, we can rewrite the loss as:
\begin{equation}
    \mathcal{L} = \frac{1}{2} \sum_\alpha s_\alpha \left( \frac{s_\alpha}{a_\alpha^\infty} a_\alpha^2(t) - 2 a_\alpha(t) s_\alpha + 1 \right).
    \label{eq:expected_loss_final_a_alpha_inf}
\end{equation}

\subsubsection{Loss at Infinity and Initial Loss Behavior}
As $t \to \infty$, $a_\alpha(t) \to a_\alpha^\infty$. The loss at infinite time (converged loss) is:
\begin{align}
    \mathcal{L}(\infty) &= \frac{1}{2} \sum_\alpha s_\alpha \left( \frac{s_\alpha}{a_\alpha^\infty} (a_\alpha^\infty)^2 - 2 a_\alpha^\infty s_\alpha + 1 \right) = \frac{1}{2} \sum_\alpha s_\alpha \left( a_\alpha^\infty s_\alpha - 2 a_\alpha^\infty s_\alpha + 1 \right) \\
    &= \frac{1}{2} \sum_\alpha s_\alpha \left( 1 - a_\alpha^\infty s_\alpha \right) = \frac{1}{2} \sum_\alpha s_\alpha \left( 1 - \frac{s_\alpha}{\frac{N+1}{N} s_\alpha + \frac{\operatorname{Tr}(S)}{N}} \right) \\
    &= \frac{1}{2} \sum_\alpha s_\alpha \left( \frac{\frac{N+1}{N} s_\alpha + \frac{\operatorname{Tr}(S)}{N} - s_\alpha}{\frac{N+1}{N} s_\alpha + \frac{\operatorname{Tr}(S)}{N}} \right) = \frac{1}{2} \sum_\alpha s_\alpha \left( \frac{\frac{1}{N} s_\alpha + \frac{\operatorname{Tr}(S)}{N}}{\frac{N+1}{N} s_\alpha + \frac{\operatorname{Tr}(S)}{N}} \right) \\
    &= \frac{1}{2} \sum_\alpha \frac{s_\alpha (\frac{1}{N} s_\alpha + \frac{\operatorname{Tr}(S)}{N})}{\frac{N+1}{N} s_\alpha + \frac{\operatorname{Tr}(S)}{N}} = \frac{1}{2N} \sum_\alpha \frac{s_\alpha (s_\alpha + \operatorname{Tr}(S))}{\frac{N+1}{N} s_\alpha + \frac{\operatorname{Tr}(S)}{N}} = \frac{1}{2} \sum_\alpha \frac{s_\alpha (s_\alpha + \operatorname{Tr}(S))}{(N+1)s_\alpha + \operatorname{Tr}(S)}.
    \label{eq:loss_infinity}
\end{align}

To understand the initial behavior of the loss, we compute the time derivative of $\mathcal{L}$ at $t=0$. Using chain rule and Equation~\eqref{eq:expected_loss_final_a_alpha}:
\begin{align*}
    \frac{d\mathcal{L}}{dt} &= \sum_\alpha \frac{\partial \mathcal{L}}{\partial a_\alpha} \frac{d a_\alpha}{dt} = \sum_\alpha \frac{1}{2} s_\alpha \left( 2 a_\alpha s_\alpha \frac{1}{a_\alpha^\infty} - 2 s_\alpha \right) \frac{d a_\alpha}{dt} = \sum_\alpha s_\alpha \left( \frac{s_\alpha a_\alpha}{a_\alpha^\infty} - s_\alpha \right) \frac{d a_\alpha}{dt} \\
    &= \sum_\alpha s_\alpha^2 \left( \frac{a_\alpha}{a_\alpha^\infty} - 1 \right) \frac{d a_\alpha}{dt}.
\end{align*}
Substituting $\frac{d a_\alpha}{dt} = \frac{2}{\tau_\alpha} a_\alpha (1 - a_\alpha s_\alpha^\infty)$ from Section~\ref{sec:dynamical_equations_conserved_quantity}:
\begin{align*}
    \frac{d\mathcal{L}}{dt} &= \sum_\alpha s_\alpha^2 \left( \frac{a_\alpha}{a_\alpha^\infty} - 1 \right) \left( \frac{2}{\tau_\alpha} a_\alpha (1 - a_\alpha s_\alpha^\infty) \right) = \sum_\alpha \frac{2 s_\alpha^2}{\tau_\alpha} \left( \frac{a_\alpha}{a_\alpha^\infty} - 1 \right) a_\alpha (1 - a_\alpha s_\alpha^\infty) \\
    &= \sum_\alpha \frac{2 s_\alpha^2}{\tau_\alpha} \left( \frac{a_\alpha}{a_\alpha^\infty} - 1 \right) a_\alpha \left( 1 - \frac{a_\alpha}{a_\alpha^\infty} \right) \qquad (\text{since } a_\alpha^\infty = 1/s_\alpha^\infty) \\
    &= - \sum_\alpha \frac{2 s_\alpha^2}{\tau_\alpha} a_\alpha \left( \frac{a_\alpha}{a_\alpha^\infty} - 1 \right)^2.
\end{align*}
At $t=0$, we have $a_\alpha = a_\alpha^0$. Thus, the initial rate of change of loss is:
\begin{equation}
    \frac{d\mathcal{L}}{dt}\Big|_{t=0} = - 2 \sum_\alpha \frac{s_\alpha^2}{\tau_\alpha} a_\alpha^0 \left( \frac{a_\alpha^0}{a_\alpha^\infty} - 1 \right)^2 = - 2 \sum_\alpha P \lambda s_\alpha^4 a_\alpha^0 \left( \frac{a_\alpha^0 - a_\alpha^\infty}{a_\alpha^\infty} \right)^2.
    \label{eq:dL_dt_t0_exact}
\end{equation}
If we assume small initial parameters $a_\alpha^0 \ll a_\alpha^\infty$, then $\left( \frac{a_\alpha^0 - a_\alpha^\infty}{a_\alpha^\infty} \right)^2 \approx 1$. In this case,
\begin{equation}
    \frac{d\mathcal{L}}{dt}\Big|_{t=0} \approx - 2 \sum_\alpha P \lambda s_\alpha^4 a_\alpha^0 = - 2 \sum_\alpha \frac{s_\alpha^2 a_\alpha^0}{\tau_\alpha}.
    \label{eq:dL_dt_t0_approx}
\end{equation}
Since $P, \lambda, s_\alpha^2, a_\alpha^0, \tau_\alpha$ are positive, the initial derivative is negative, indicating that the loss function initially decreases. The rate of decrease is proportional to the initial parameter values $a_\alpha^0$ and the eigenvalues $s_\alpha^2$ of the covariance matrix.

\end{document}